# The Information-Theoretic Imperative and Compression Efficiency: Why Brains and Deep Networks Converge

**Christian Dittrich, PhD**[1] and **Jennifer Kinne**[2]

[1] Sciquoia AG, Grellingerstrasse 77, 4052 Basel, Switzerland
[2] Harvard University, Faculty of Arts & Sciences, 1350 Massachusetts Ave, Suite 350, Cambridge, MA 02138, USA

**Corresponding author:** Jennifer Flygare Kinne (flygare@fas.harvard.edu)

## Abstract

Why do brains and deep networks converge on similar representations? Task-optimized artificial neural networks quantitatively predict primate ventral stream responses despite radically different substrates and optimization dynamics. This convergence demands explanation beyond shared natural image statistics or task structure alone.

The Compression Efficiency Principle (CEP) specifies the selection mechanism: representations exploiting unstable correlations pay a growing "exception tax" (approximately linear excess codelength under shortcut-flipping shifts), while representations encoding shift-stable invariants amortize this cost. When environments provide intervention-rich shifts and exhibit approximately modular causal structure, these invariants align with causal mechanisms.

The framework offers a unified lens on three biological signatures—steep metabolic constraints on neural signaling, high coding efficiency in early sensory pathways, and hierarchical tolerance in the ventral stream—and connects them to parallel phenomena in deep learning: scaling frontiers, shortcut failures under distribution shift, and the role of augmentation in enforcing invariances.

Distinctive predictions follow: a crossover threshold beyond which invariant representations dominate, and systematic coupling between compression efficiency and out-of-distribution robustness—testable across substrates. Predicted divergences (sparse biological signaling versus dense overparameterization) arise from different resource constraints on a shared trade-off topology.

The convergence is not a coincidence. It is evidence for a substrate-independent basin shaped by predictive compression under shift.

# 1. Introduction: the convergence problem

In primate vision, cortical activity in the ventral stream is not an arbitrary encoding of images; it is structured in a way that supports robust object recognition across nuisance variation [1, 2, 3]. What is unexpected is that engineered deep networks—trained to optimize predictive performance—produce internal representations that quantitatively predict these neural responses. In a canonical result, performance-optimized convolutional neural networks (CNNs) account for a substantial fraction of the explainable variance in macaque inferior temporal (IT) responses, with a coarse hierarchy-to-hierarchy correspondence in which earlier network stages best predict earlier visual areas and later stages best predict IT [4,5]. Subsequent work has systematized this alignment into benchmarkable metrics (e.g., Brain-Score [3]), showing both that (i) alignment is robust enough to be tracked quantitatively and (ii) it is not "solved": models continue to climb the alignment frontier as architectures and training regimes evolve [3].

This alignment is not isolated. Across biological and artificial neural networks, a recurring set of representational motifs appears: sparse and selective codes [6], hierarchical abstraction [1], approximate invariances to nuisance transformations [1], and prediction-oriented processing [7] (Table 1). Artificial networks additionally exhibit systematic phenomena—scaling behavior [8], emergent adaptation, and characteristic failure modes under distribution shift [9]—that mirror, in a different register, long-studied phenomena in neuroscience such as efficient coding [6] and predictive processing [7]. Yet a unified physical explanation for why these motifs recur across substrates remains incomplete.

**Table 1. Convergence inventory (representative quantitative anchors)**

This table is illustrative, not exhaustive; values depend on task, dataset, estimator, and recording regime.

| Dimension | Biological neural systems | Artificial neural systems | Representative quantitative anchor |
|---|---|---|---|
| Brain–model representational alignment | Ventral stream responses structured by task demands | Task-trained vision models yield predictive representations | CNN features predict IT responses; alignment trackable as a benchmark [3,4,5] |
| Efficient coding / frontier-seeking under constraints | Sensory systems approach coding limits under metabolic constraints [10,11] | Models approach performance–resource frontiers under compute/data constraints [8] | High coding efficiency in sensory pathways [6]; reproducible performance–resource frontiers (Section 4) |
| Hierarchy + invariance | V1→V4→IT progression supports selectivity+tolerance | Deep hierarchies support invariance and abstraction | Increasing tolerance/selectivity along ventral stream [1, 12] with analogues under engineered shift families (Section 4) |
| Failure under distribution shift | Illusions / context effects: locally efficient codes misaligned with specific stimuli | Shortcut learning / spurious features: in-distribution success, OOD brittleness | Both exhibit off-frontier behavior under atypical/adversarial inputs (formalized via regret under shift in Sections 2 and 6) |
| Resource trade-offs (informative divergence) | Strong metabolic/wiring constraints encourage sparse signaling and local learning rules | Compute abundance permits dense, overparameterized training; different trade-offs | Different operating points on a shared rate–distortion/resource frontier (Section 5) |

## 1.1 The central mystery: why should these systems converge?

The convergence is puzzling because the generative histories of the systems are maximally different:

● **Biological neural networks** are shaped over evolutionary time by differential persistence and reproduction [9], implemented in biochemical substrates with strict metabolic budgets [13, 14] and local plasticity rules.

● **Artificial neural networks** are shaped on computational timescales by explicit optimization objectives (typically predictive loss minimization), implemented in digital substrates and trained by optimization procedures (e.g., backpropagation) that have no direct biological analogue (15, 16).

A skeptic might argue this is not mysterious: both brains and models are exposed to similar natural image statistics and solve similar recognition tasks, so "of course" they learn similar features. But this misses the sharper point. The optimization paths and constraints are profoundly different (selection + development + local plasticity under metabolic constraints versus gradient-based optimization with abundant compute). The fact that these distinct processes nonetheless yield measurable representational alignment suggests that the solution is not a fragile accident of one particular training procedure, but is being selected by a shared constraint.

**Null hypothesis (data/task/architecture sufficiency) and how to distinguish it.**

A strong null explanation is that brain–model alignment is primarily a consequence of (i) shared natural image statistics [17], (ii) shared task structure (e.g., categorization [4]), and (iii) architectural inductive biases (e.g., convolution), rather than a substrate-independent constraint. Indeed, task choice and training objective matter: models trained to optimize behaviorally relevant objectives tend to predict ventral-stream responses better than models that do not [18]. ITI/CEP does not deny task/data dependence; it predicts how it matters: alignment and robustness should increase specifically when training exposure approximates the organism's shift family ε (e.g., through augmentation or interaction) and when constraints bind so that patching becomes costly. This yields discriminating tests: holding data and architecture fixed, systematically vary the experienced shift family ε and measure both (i) brain alignment (Brain-Score-style metrics) and (ii) shift-regret scaling (Section 6). A "data/architecture sufficiency" null predicts alignment will be largely explained by in-distribution task accuracy alone; ITI/CEP predicts additional variance explained by shift-family match and by reductions in cross-environment excess codelength.

This motivates the central question of this review: **What substrate-independent constraint channels both evolved and engineered learning systems toward similar representational solutions?**

To ground this inquiry, we anchor our framework in resolving three specific biological mysteries:

**The Brain's Metabolic Budget**: The human brain consumes ~20% of resting metabolic energy despite comprising ~2% of body mass [13]. Why does natural selection sustain this cost rather than selecting for cheaper, reactive strategies?

**Sensory Coding Efficiency**: Early sensory pathways, such as blowfly motion-sensitive neurons, achieve up to ~90% of theoretical channel capacity under natural conditions [10]. Why such extreme, frontier-seeking optimization in a "simple" organism?

**The Invariance Explosion (V1 → IT)**: The ventral stream transforms high-dimensional retinal input into low-dimensional object representations tolerant to extreme nuisance variation (position, scale, rotation) [1, 12]. How does this hierarchy reliably discover the right invariances?

The framework developed in Sections 2–4 resolves all three through a single physical constraint; Section 5 closes the loop.

## 1.2 What existing frameworks explain—and what they leave underspecified

Several powerful frameworks already connect intelligence-like behavior to prediction and compression:

● **Efficient coding** explains why sensory systems reduce redundancy and approach information-theoretic limits under metabolic constraints [6,10].

● The **Free Energy Principle (FEP)** casts biological self-organization as minimizing a bound on surprisal through perception and action [7,19].

● **Minimum Description Length (MDL)** formalizes inference as compression and operationalizes parsimony via total code length [20,21].

● **Information Bottleneck (IB)** formalizes the compression–prediction trade-off for representations [22], while also motivating rate–distortion perspectives on representation learning [23].

● **Predictive coding** supplies an implementational story for hierarchical prediction-error processing in cortex [7,24].

● **Compression progress** accounts for curiosity-driven learning dynamics via improvements in compressibility [25, 26].

These accounts converge on the importance of prediction and compression. The remaining gap—central for explaining brain–AI convergence—is mechanistic under distribution shift:

**Why does the pressure for efficient predictive compression select representations that track shift-stable structure, rather than representations that exploit local correlations that are compressible but brittle?**

Compression can generate compact fictions. The convergence problem demands an account of why, in the relevant regime, efficient compression is systematically constrained toward structure that remains predictive across contexts.

## 1.3 Thesis overview: ITI → CEP as a constraint channel

The principles underlying this synthesis are not new in isolation. That persistence requires prediction is recognized across cybernetics [27], thermodynamics [14], and evolutionary theory [9]. That compression discovers structure is central to MDL [20,21], efficient coding [6,10], and compression progress [25,26]. That invariances support generalization is the core insight of causal representation learning [28,29].

What has been missing is an account of why these principles co-occur—why systems that persist under uncertainty tend to exhibit efficient coding, predictive compression, and invariant representations simultaneously, whether evolved or engineered.

Our answer is **selection**. Under finite resources and recurring distribution shift, representational strategies face differential survival. Shortcuts that exploit shift-unstable correlations incur growing costs as environments diversify; representations encoding shift-stable structure amortize these costs. The constraint channel we describe—from persistence through prediction and compression to shift-stable invariants—is not a new theory but a unifying lens that explains why independent frameworks converge on overlapping predictions and why biological and artificial systems converge on similar representations [3].

We organize this synthesis under two labels. The **Information-Theoretic Imperative (ITI)** names the constraint: for systems that persist through prediction (active persisters, as defined in Section 2.1), persistence under uncertainty and finite resources creates strong selection pressure toward predictive compression.The **Compression Efficiency Principle (CEP)** names the selection mechanism: under recurring shift, efficient compression penalizes shortcuts via excess codelength and favors shift-stable invariants. When environments provide intervention-rich shifts and exhibit approximately modular causal structure (conditions detailed in Section 2.3), these invariants align with causal mechanisms.

The novel contribution is not the components but the constraint channel linking them—and the quantitative predictions this channel generates, particularly the crossover threshold E* at which invariant representations become selectively dominant.

This yields the constraint channel:

**[Persistence ⇒ Prediction ⇒ Compression ⇒ Efficiency under budgets ⇒ Shift-stable invariants]**

Under specific environmental conditions (intervention-rich shifts + approximately modular causal structure), these shift-stable invariants correspond to causal mechanisms (Section 2.3).

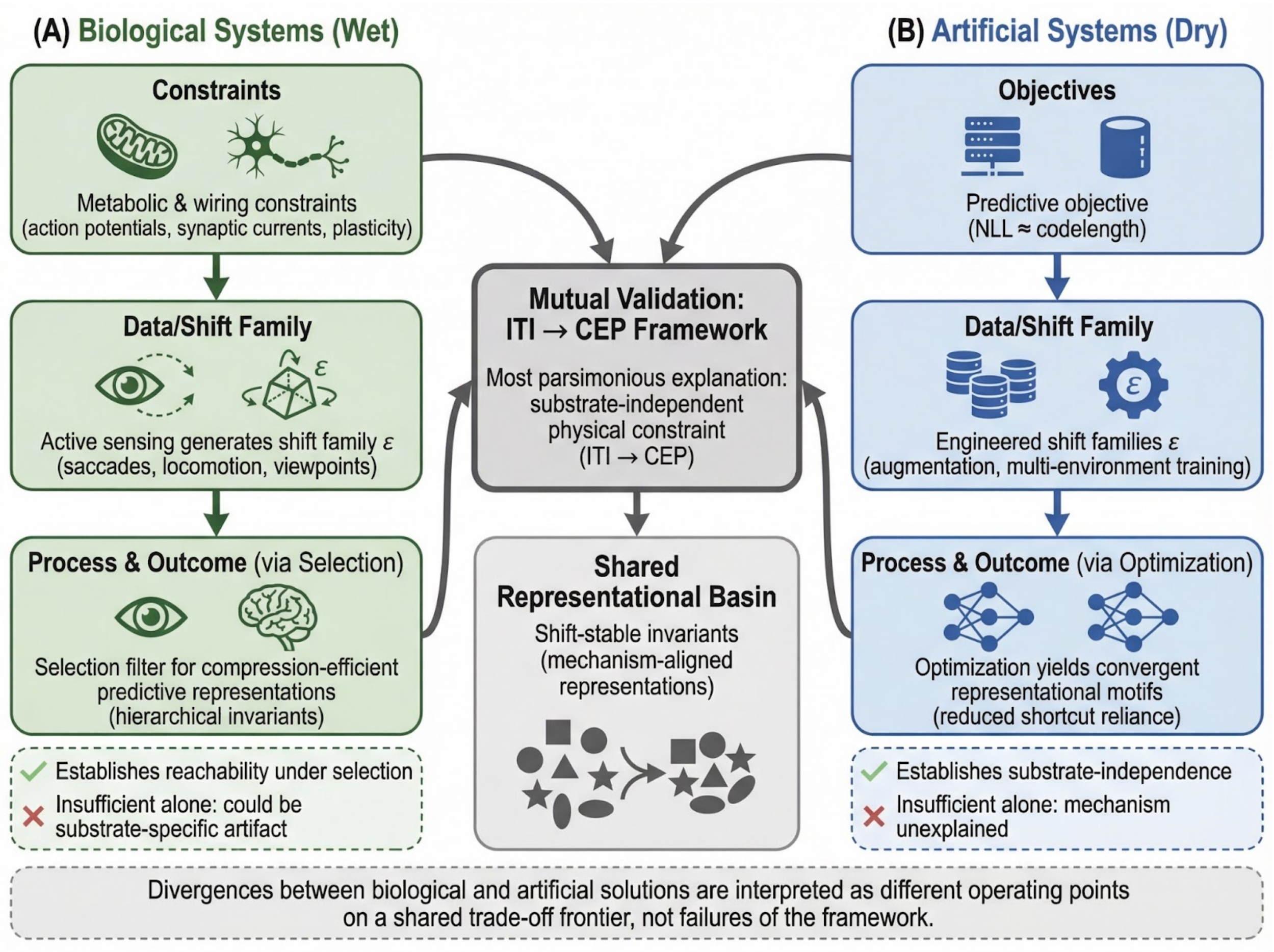

**Figure 1. Dual-substrate convergence as evidence for a shared physical constraint (A)** In biological systems, evolutionary selection under metabolic and wiring constraints, combined with active sensing that generates a natural shift family ε, favors compression-efficient predictive representations. This establishes reachability under selection but does not rule out substrate-specific explanations. **(B)** In artificial systems, optimization under resource budgets with predictive objectives (NLL ≈ codelength) and engineered shift families ε yields convergent representational motifs, including reduced shortcut reliance. This establishes substrate-independence but does not identify the constraining mechanism. (Center) Their conjunction supports a shared constraint channel (ITI → CEP): efficient prediction under finite resources and distribution shift channels both substrates toward a common basin of shift-stable, mechanism-aligned representations. Divergences are interpreted as different operating points on a shared trade-off frontier, not failures of the framework.

## 1.4 Central quantities and the shift-based operationalization

To connect biology and AI in comparable terms, we adopt an operational stance: the primary measurable objects are **predictive codelength** and **excess codelength (regret)** under a shift family (ε) [30].

**1) Predictive codelength under shift (primary).** When models are trained and evaluated using a normalized likelihood, negative log-likelihood is predictive codelength [20,21]. Under a shift family (ε), changes in codelength constitute excess codelength—the operational "exception tax" of CEP.

**2) Frontier-relative efficiency (primary).** We operationalize "compression efficiency" as improvements in predictive codelength per unit resource (parameters/compute/energy), evaluated within a fixed task and shift family. This is the quantity used in our falsifiers and protocols (Section 6).

**Terminology note.** Invariance: throughout this manuscript, “invariant” denotes a representation Z for which the predictive conditional $P_e(Y|Z)$ remains approximately stable across environments $e \in \varepsilon$. The representation values themselves may change; what is invariant is the conditional distribution. This is consistent with standard usage in causal representation learning [28,29]. When we discuss “hierarchical invariance” in the ventral stream (Section 3.3), we refer to the neuroscience concept of tolerance—the progressive robustness of neural responses to nuisance transformations [12]. These concepts are related but distinct: perceptual tolerance is a candidate mechanism for achieving conditional-distribution stability under a natural shift family ε, but tolerance to transformations does not by itself guarantee that $P_e(Y|Z)$ is shift-stable. Where the distinction matters, we specify which sense is intended.

**Conceptual note.** Information Bottleneck [22] and rate–distortion [23] provide useful conceptual scaffolding for compression–prediction trade-offs (reviewed in Section 2.4.3), but our empirical program does not rely on estimating high-dimensional mutual information.

## 1.5 Mutual validation: why convergence is evidence for a physical constraint

The convergence provides a distinctive form of support unavailable to biology-only or AI-only accounts:

- **Biology demonstrates reachability under selection.** Over evolutionary time, systems subject to metabolic and ecological constraints converge toward efficient predictive representations [10,11].
- **AI demonstrates substrate-independence.** With different substrates and optimization dynamics, systems optimizing predictive codelength under constraints rediscover similar representational motifs and, in key cases, align quantitatively with neural responses [4,5].

Together, this cross-substrate convergence supports the claim that compression efficiency under selection for predictive accuracy is a physical constraint shaping representational strategy.

A further benefit of the constraint view is that it predicts **informative divergences**. For example, strong metabolic and wiring constraints in brains encourage sparse signaling [31,32] and local learning rules, whereas compute-abundant artificial systems can sustain dense overparameterized representations [8]; these differences can be interpreted as different operating points on a shared rate–distortion/resource frontier (Section 5).

### 1.6 Roadmap and falsifiability teaser

Section 2 formalizes ITI and CEP and positions them relative to FEP, MDL, IB, efficient coding, and compression progress, with explicit assumptions and falsification criteria. Sections 3–4 marshal biological and artificial evidence symmetrically, emphasizing both convergences and informative divergences as different trade-offs. Section 5 synthesizes the ITI→CEP channel as the proposed resolution of the convergence puzzle. Section 6 derives quantitative predictions using regret/codelength and frontier-relative efficiency proxies across substrates.

The framework is falsifiable in a strong sense: if high proxy-efficiency systems systematically fail to exhibit improved out-of-distribution robustness under controlled shifts [9], or if invariant/mechanism representations do not reduce cross-environment regret relative to shortcut strategies, the CEP mechanism must be revised or rejected (Sections 2 and 6).

**Summary.** The problem is not that brains and deep networks share surface similarities; it is that they exhibit quantitative representational alignment [3,4] despite different substrates and optimization dynamics [1]. We argue that this is best explained by a shared constraint channel: selection for persistence yields predictive compression under finite resources (ITI), and under distribution shift, compression efficiency favors shift-stable invariants, thereby producing convergent representational solutions across biology and engineered AI. When environments provide intervention-rich shifts and exhibit modular causal structure, these invariants correspond to causal mechanisms (detailed in Section 2.3).

# 2. Theoretical framework: ITI → CEP under shift

This section presents the theoretical core in a constraint-based form. The key move is to separate why predictive compression is unavoidable for active persisters (ITI) from what representational strategies remain efficient under distribution shift (CEP). Section 2.3 then specifies the conditional relationship between shift-stable invariants and causal mechanisms.

## 2.0 Setup: prediction across environments, with explicit variables

We consider systems that learn representations for prediction across a family of environments. Let

● $e \in \varepsilon$ index environments (contexts, domains, regimes, or interventions). ● Each e induces a joint distribution $P_e(X,Y)$ over observations X and prediction targets Y. ● A representation Z is produced by an encoder $q_\varphi(z|x)$, and predictions are made by $p_\theta(y|z)$.

What is Y? We use three regimes, each explicitly scoped:

1. Passive supervised: Y is a label or latent property of X (e.g., object identity).
2. Predictive / time-series: $Y = X_{t+1:t+k}$ is a future segment predicted from $X_{1:t}$.
3. Agentic: future outcomes depend on actions and latent state. We index time explicitly: a representation $Z_t$ is formed from past observations $X_{\leq t}$, an action $A_t \sim \pi(\cdot|Z_t)$ is then taken, and the next outcome $Y_{t+1}$ is realized from environmental dynamics. In this manuscript, our core CEP claims and protocols are formulated in the passive/shifted-environment regime; agentic settings enter through the way action generates richer shift families $\varepsilon$ (Sections 3.3, 4.3, 6.5).

**2.1 ITI: persistence constrains systems toward predictive compression**

Information-Theoretic Imperative (ITI) (constraint statement).

For active persisters—systems that maintain organized states in uncertain environments via ongoing regulation (excluding systems persisting through replication speed, thermodynamic stability, or other non-predictive mechanisms)—selection for persistence strongly favors strategies that reduce expected surprisal by building predictive models [7] under finite resources (energy, memory, time).Under bounded resources, prediction requires compression: replacing growing raw histories with compact internal structure that preserves predictive relevance [33,34].

ITI is compatible with multiple traditions:

- Cybernetics / requisite variety. Maintaining viability against disturbances requires sufficient informational coupling between sensed disturbances and compensatory actions [27, 35].
- Information theory. Long-run prediction without storing the entire history requires exploiting regularities—i.e., compressibility—in the sensory stream [33,35].
- Neuroenergetics / efficient coding. Biological substrates impose severe resource constraints [10,13,31,32], making efficient representation [6,10,17] a prerequisite for scalable prediction.

Operationally, a useful handle is expected predictive codelength (log-loss). For a predictor $p_\theta(y|z)$, define cumulative predictive loss under an environment sequence $(e_t)$:

$$L_T(\varphi,\theta) \equiv \Sigma_{t=1}^T E_{(X_t,Y_t)\sim P_{e_t}}[-\log p_\theta(Y_t|Z_t)],\ Z_t \sim q_\varphi(\cdot|X_t).$$

ITI says: among systems under persistent selection pressure, strategies that achieve low long-run $L_T$ per unit resource are favored.

## 2.2 CEP: efficient predictive compression selects shift-stable invariants (via regret/excess codelength)

### 2.2.1 The CEP object: excess codelength (regret) under shift

To make "exception accumulation" mathematically grounded, CEP uses a standard object shared by MDL and online learning:

● Codelength (MDL / universal coding): total description length of model + data given model [22,23] ● Regret (online prediction): excess cumulative log-loss relative to a reference (e.g., best fixed model in a class) [37,38].

Let M denote a model class (or representation–predictor class). The regret of a sequential predictor $\hat{m}_t$ relative to the best fixed $m \in M$ for the realized sequence is:

$$\text{Regret}_T(M) \equiv \Sigma_{t=1}^T -\log p_{\hat{m}_{t-1}}(y_t|x_t) - \min_{m\in M} \Sigma_{t=1}^T -\log p_m(y_t|x_t).$$

The "reference" can be chosen to match the scientific question (e.g., best fixed invariant predictor in $M_{inv}$; best context-oblivious shortcut predictor in $M_{short}$; or best environment-aware predictor in a class allowed to condition on e when e is observable).

Under standard regularity assumptions, even well-specified models incur sublinear regret due to estimation uncertainty [32,60]; in parametric settings this is often $\tilde{O}(\log T)$, whereas in nonparametric settings it can be $\tilde{O}(\sqrt{T})$ or other sublinear forms. CEP's distinguishing claim is comparative under a fixed resource budget: shortcut strategies either incur linear excess codelength (see Appendix A.8), or must pay model-cost that grows with the number of contexts to represent patch rules. Background on universal prediction under nonstationarity is provided by [30,40].

Crucially, the independent variable is environmental diversity, which can be indexed in different but related ways:

● Online time (T) when shifts recur with nonzero rate in a nonstationary stream, ● Number of distinct environments (E) encountered (domains/contexts), when evaluating transfer across $\{P_e\}_{e \in \varepsilon}$, ● or an effective complexity measure of the shift family.

### 2.2.2 CEP statement (defensible, conditional)

Compression Efficiency Principle (CEP).

Fix a family of environments ε and a resource budget (constraining representational capacity, update cost, or total description length). Among feasible predictive strategies, those that minimize long-run predictive codelength across ε preferentially encode invariants—structure for which the conditional $P_e(Y|Z)$ remains stable (or approximately stable) across $e \in \varepsilon$. Strategies that rely on correlations whose predictive value changes across e pay sustained excess codelength (regret) because they require context-specific corrections. The relationship between these shift-stable invariants and causal structure is addressed in Section 2.3.

Rate clarification. CEP's core empirical signature is the separation between linear patch-driven growth and sublinear estimation-driven growth—not a specific sublinear exponent. In parametric well-specified settings, standard online-learning theory yields $\tilde{O}(\log T)$ regret for invariant strategies [37,38]; in nonparametric or misspecified settings, rates such as $\tilde{O}(\sqrt{T})$ are common [37]. The key testable claim is that the between-class gap $G_T$ between shortcut and invariant strategies grows as $\Omega(T)$—i.e., the shortcut penalty is linear—while the invariant class remains sublinear at whatever rate the effective model complexity permits. This separation, rather than any particular sublinear exponent, is what Section 6 operationalizes.

### 2.2.3 A minimal toy separation (why shortcuts pay linear regret)

Let the environment switch among contexts $e \in \{0,1\}$. Observations contain two features $X=(X_1,X_2)$ and a target $Y \in \{0,1\}$. Suppose:

● a stable mechanism feature: $Y = X_2$ in both contexts, and ● a shortcut feature whose correlation flips: ○ if e=0: $X_1 = Y$, ○ if e=1: $X_1 = 1-Y$.

Assume contexts recur with a nonzero (ergodic) rate and are not directly observed. Consider:

● Shortcut class $M_{short}$: predictors that use $X_1$ and do not represent e, ● Invariant class $M_{inv}$: predictors that use $X_2$.

Whenever the context differs from what the shortcut implicitly "expects," the shortcut incurs a nonzero per-step expected log-loss gap bounded below by a constant. Therefore, under recurrent context switching with nonzero rate,

$$\mathrm{Regret}_T(M_{short}) = \Omega(T).$$

By contrast, the invariant predictor class is well-specified in both contexts and pays only estimation regret,

$$\mathrm{Regret}_T(M_{inv}) = \tilde{O}(\log T),$$

under standard parametric assumptions [60] (or more generally, sublinear rates determined by the complexity of the invariant class). Appendix A.8 (Proposition A.2) formalizes the complementary case in which a system does patch contexts: if low loss requires environment-specific corrections, model description length grows at least linearly in the number of distinct environments (E), yielding a crossover E* beyond which invariant representations are preferred under MDL-style total codelength minimization.

## 2.3 When invariants become causal: interventional diversity + modular mechanisms

### 2.3.0 Epistemic status of the causality bridge

CEP is a selection principle, not a causal oracle. It does not explain where causal structure comes from; it explains why, when causal structure exists and is accessible through intervention-rich shifts, mechanism-aligned representations are the only compression-efficient equilibrium under resource constraints.

The bridge to causality rests on an empirical premise about environmental structure: that many natural environments generate intervention-rich, approximately modular shifts [28,29]. CEP does not derive this structure; it explains why, when such structure is present, mechanism-aligned representations are uniquely compression-efficient.

This conditioning has three important implications:

1. CEP explains selection, not genesis. The framework does not claim that compression pressure alone creates causal structure. It claims that when environments contain causal structure organized into approximately modular mechanisms, and when that structure is accessible through interventions or regime changes, then representations aligned to those mechanisms will be favored by the compression-efficiency criterion.
2. ICM-richness is an environmental property, not a theorem. Whether a shift family $\varepsilon$ is "ICM-rich" (contains interventions that perturb mechanisms modularly) is an empirical question about the world, not a derivable consequence of information theory. Natural environments may or may not exhibit this property, and the degree of ICM-richness can vary.

3. Complementarity with causal discovery. This makes ITI/CEP complementary to causal discovery methods rather than competing with them. Causal discovery methods aim to identify causal structure from data; ITI/CEP explains why, once accessible, causal structure is compression-efficient and thus subject to selection pressure. The frameworks address different questions.

### 2.3.1 Invariants are not automatically causal

Box 2.1. Invariants are not automatically causal: counterexamples and scope conditions

Before stating the causality bridge formally, we must establish what it does not claim. CEP's first-order selection target under shift is invariance: features or representations Z for which the predictive mapping $P_e(Y|Z)$ remains approximately stable across $e \in \varepsilon$. This is necessary for low cross-environment excess codelength, but it is not sufficient for causal identification [28,29]. The following examples clarify the scope conditions.

1. Symmetry and group invariants (non-mechanistic). Many environments exhibit stable symmetries (e.g., translation or rotation in images). Representations that factor group orbits can be highly shift-stable and compressive without uniquely corresponding to causal variables or mechanisms.
2. Stable confounding and selection effects. A latent confounder U can induce a stable relationship between Z and Y across observed environments even when Z is not a cause of Y [29]. If the shift family ε does not perturb the confounding pathway (or does so in a correlated way), the spurious association can remain invariant and thus "survive" CEP selection.
3. Dataset/measurement invariants (instrumental artifacts). Persistent properties of the measurement pipeline (camera/codec artifacts, annotation conventions) or dataset bias can be invariant across domains and highly predictive in the dataset, yet unrelated to the underlying generative mechanisms of interest [9].
4. Algorithmic/structural invariants without intervention meaning. Periodicities or parity-like features can be compressible and stable across many shifts while lacking the modular "intervention semantics" associated with mechanisms. Such invariants can reduce codelength under observational shifts without supporting localized updates under targeted interventions.

What CEP does (and does not) claim.

● CEP does claim: under a specified shift family ε, compression efficiency penalizes shift-unstable correlations via excess codelength and favors shift-stable invariants.

● CEP does not claim: invariance alone is sufficient for causal identification or uniquely determines causal structure [28,29].

● CEP does claim (conditional bridge): under ICM-rich shift families and approximate mechanism modularity, mechanism-aligned invariants are favored (ICM/modularity premise: [28,29]) because they localize updates and reduce cross-environment codelength under resource constraints.

This conditionality clarifies why invariance-based objectives (e.g., IRM [37]) may fail to recover mechanisms when $\varepsilon$ is not ICM-rich, when multiple invariant predictors exist, or when invariants correspond to artifacts rather than mechanisms [37]. ITI/CEP does not claim to eliminate these failure modes; it predicts when they arise and supplies a distinct selection variable—cross-environment codelength under resource constraints—that can be tested experimentally (Section 6).

Practical diagnostic (protocol-level). To test whether an invariant is mechanism-aligned, evaluate it under targeted interventions that independently perturb nuisance correlations and candidate mechanisms. Mechanism-aligned invariants preserve predictive structure with localized updating; spurious invariants do not.

### 2.3.2 Independent Causal Mechanisms (ICM) / modularity as the bridge

The core bridge from invariance to causality relies on two empirical premises about the structure of many natural environments:

Premise 1: Independent Causal Mechanisms (ICM). Natural systems often exhibit causal structure that can be decomposed into approximately autonomous modules (mechanisms), such that intervening on one mechanism does not systematically alter the parameters of other mechanisms [28,29]. For example, in vision: an object's shape-generating process is approximately independent of the illumination process [28]; changing the lighting does not restructure the geometry of objects.

Premise 2: Interventional accessibility. Many natural environments—particularly those experienced by organisms capable of action, or experimental systems where controlled manipulations are possible—generate shifts that approximate targeted interventions, perturbing one or a few mechanisms while leaving others approximately invariant.

Under these premises, mechanism-aligned representations become uniquely compression-efficient because they localize the adaptation required when environments change. When a shift perturbs only the lighting mechanism, a representation that separately encodes "shape" and "illumination effects" requires updating only the illumination component; an entangled representation encoding pixel-level correlates must undergo broader reconfiguration.

How to identify ICM-richness operationally. Determining whether a shift family ε is ICM-rich requires one of the following:

1. Explicit experimental control over intervention targets (e.g., laboratory settings where you know you're manipulating lighting vs. shape independently).
2. Natural experiments where interventions can be identified from domain knowledge (e.g., policy changes that affect one mechanism, evolutionary regime shifts, developmental transitions).
3. Posterior validation that learned representations support localized adaptation: when one environmental factor changes, only a small subset of representational parameters requires updating. This is a testable signature rather than a prior definition (Section 6).

Crucially: Passive observation of a stationary distribution is generally insufficient to identify ICM-richness. The framework predicts that when ICM-rich shifts are available—whether through active exploration, experimental intervention, or natural regime changes—mechanism-aligned representations will be selected by compression efficiency. When such shifts are not available, CEP still selects invariants, but those invariants may correspond to stable confounders, symmetries, or other non-causal regularities (see Section 2.3.1).

In biological substrates, this localization reduces not only stored description length but also the physical cost of adaptation, since synaptic remodeling and plasticity-driven reconfiguration are metabolically expensive; entangled representations require broader reconfiguration to track the same environmental change.

This perspective aligns with invariance-based approaches in machine learning (e.g., Invariant Risk Minimization [37]), which can be interpreted as attempts to approximate mechanism-stable predictors across environments. We treat these as complementary operationalizations of the same underlying selection pressure rather than as definitive solutions to causal identification.

### 2.3.3 Conditional causality bridge

Causality bridge (conditional and falsifiable).

Efficient predictive compression under shift selects shift-stable invariants (CEP core claim, Section 2.2).

When the following conditions hold: (i) The shift family ε is ICM-rich: it contains interventions or regime changes that approximately perturb each mechanism module independently while leaving other modules invariant (or only weakly affected), and (ii) The underlying data-generating process exhibits approximate mechanism modularity: the causal structure can be decomposed into approximately autonomous modules whose parameters can be changed independently,

Then: the invariants that minimize cross-environment regret correspond to causal-generative mechanisms in the operational sense of supporting intervention-invariant prediction and localized updating [28,29,37].

Why this follows from CEP. This is not a separate claim but a consequence of compression efficiency under the specified conditions. Mechanisms are precisely the representational units that: ● Remain predictively valid across the interventional shifts in ε (shift-stable) ● Require only local updates when individual environmental factors change (compression-efficient) ● Minimize cross-environment codelength by avoiding the accumulation of environment-specific patches

Therefore, when environments are ICM-rich and mechanisms are modular, mechanism-aligned representations are the unique compression-efficient solution. Representations that entangle mechanisms or rely on non-mechanistic shortcuts will incur either linear regret (for shortcuts that flip across interventions) or growing model complexity (for patch-based strategies), as formalized in Section 2.2 and Appendix A.8.

Falsifiability. This bridge is empirically testable: under controlled ICM-rich interventional regimes with known modular structure, compression-efficient representations should converge to mechanism-aligned encodings and exhibit localized adaptation. If they instead converge to non-mechanistic invariants requiring comparable global patching, the stated conditions are insufficient or the bridge fails (Falsifier 3, Section 2.4).

Scope and limitations. This causality bridge: ● Does not claim that compression alone identifies causal structure from passive observation ● Does not eliminate the need for intervention, domain knowledge, or experimental design in causal discovery ● Does claim that when ICM-rich shifts are available (through action, experiment, or natural regime changes), compression efficiency provides selection pressure toward mechanism-aligned representations ● Is strongest in domains where modularity assumptions are justified (many physical and biological systems) and weakest where causal structure is highly entangled or shifts are purely observational

### 2.4 Relationship to existing frameworks (balanced, with controversies)

ITI/CEP is intended as a synthesis: it inherits correct pieces of neighboring frameworks while adding a specific missing link—why efficiency under shift selects invariants and thereby explains brain–AI convergence.

#### 2.4.1 Free Energy Principle (FEP): objective vs constraint vs mechanism

FEP provides a unifying process-level objective for biological systems: minimize variational free energy as a tractable bound on surprisal [7,19,38]. Predictive coding [7,26] implements this principle through hierarchical prediction error minimization [24].

● FEP (process/objective): what biological systems minimize and how perception/action can be cast as variational inference. ● ITI (constraint): why prediction/compression is forced by persistence under finite resources for active persisters. ● CEP (mechanism under shift): why efficiency under nonstationarity/interventions penalizes shortcuts via regret and favors invariants/mechanisms.

A distinctive empirical contribution of CEP (not entailed by FEP stated only at the objective level) is the scaling structure under controlled variation in shift richness: (i) a linear-versus-sublinear separation in cross-environment excess codelength for shortcut/patch strategies versus invariant-capable strategies (Sections 2.2 and 6.3), and (ii) a crossover $E^*$ determined by the ratio of invariant structural cost to patching cost (Appendix A.8). FEP specifies a broad objective (minimizing a bound on surprisal) that can be satisfied by multiple representational solutions; ITI/CEP predicts which solutions are selectively stable under binding budgets and how stability changes as $\varepsilon$ becomes richer.

#### 2.4.2 MDL: from parsimony to cross-environment codelength minimization

MDL formalizes the model selection principle "prefer shorter total descriptions" [20,21]. CEP can be read as extending MDL from stationary settings to families of environments: compressions that look good in one environment can fail across $\varepsilon$ by incurring regret from accumulating patches.

#### 2.4.3 Information Bottleneck (IB): foundational trade-off, careful about deep-learning claims

IB provides a formal trade-off between compression and predictive relevance [22,34]. We use it as:

● a scaffold for defining representational efficiency when an appropriate dependency structure holds, and ● a bridge to rate–distortion thinking.

We do not treat "IB explains deep learning training dynamics" as settled. There is substantial debate about whether and when deep networks exhibit the compression behavior posited in some IB interpretations, and about the reliability of high-dimensional mutual-information estimates [43,44,45,46,47,48]. For ITI/CEP, the crucial empirical object is therefore not "layerwise MI compression" per se, but predictive codelength / regret under shift, which is directly measurable.

#### 2.4.4 Schmidhuber's compression progress: learning dynamics vs equilibrium efficiency under shift

Compression progress explains curiosity-driven behavior via improvements in compressibility [25,26]. ITI/CEP is complementary: it focuses on the selection of stable representational solutions under resource constraints and environmental shift, not on the intrinsic reward signals driving exploration.

### 2.5 Falsification criteria (global, quantitative)

ITI/CEP makes testable claims across substrates.

Falsifier 1: efficiency–robustness decoupling. If proxy measures of compression efficiency (frontier-relative predictive codelength per unit representational budget) are consistently uncorrelated with out-of-distribution robustness across controlled shift suites (e.g., corr<0.2 with adequate statistical power), CEP's core claim fails.

Falsifier 2: no regret separation under "shortcut-flipping" shifts. In environments designed so that shortcuts change across contexts while invariant structure remains stable, CEP predicts:

● shortcut strategies: Regret = $\Omega(E)$ (or $\Omega(T)$ when shifts recur), ● invariant-capable strategies: Regret sublinear (often $\tilde{O}(\log T)$ under parametric assumptions) [30,37,38,40]. ● If controlled experiments fail to exhibit this separation, the exception-tax mechanism is wrong or incomplete.

Falsifier 3: interventions do not favor modular/mechanism representations. Under ICM-rich interventional regimes with known modular structure, CEP predicts that mechanism-aligned representations reduce cross-environment codelength and localize updates. If compression-efficient predictors converge to non-mechanistic invariants requiring comparable global patching, the CEP→causality bridge fails.

Falsifier 4: persistence without predictive compression under resource constraints. If active persisters in uncertain environments under resource constraints can be shown to maintain viability without predictive compression beyond trivial reflex arcs, ITI must be revised. (We tighten "active persister" scope in Section 3.)

**2.6 Section 2 summary**

1. ITI: persistence under uncertainty constrains systems toward prediction and predictive compression under finite resources.
2. CEP: under shift, shortcuts pay linear regret in environmental diversity; invariants pay sublinear regret, staying near the frontier.
3. Causality bridge (conditional): When environments provide intervention-rich shifts and exhibit approximately modular causal structure, the shift-stable invariants selected by CEP correspond to causal mechanisms. This is an empirical premise about environmental structure, not a derivation from information theory (Section 2.3).
4. The synthesis is falsifiable, and it is positioned as complementary to FEP/MDL/IB/Schmidhuber rather than a replacement.

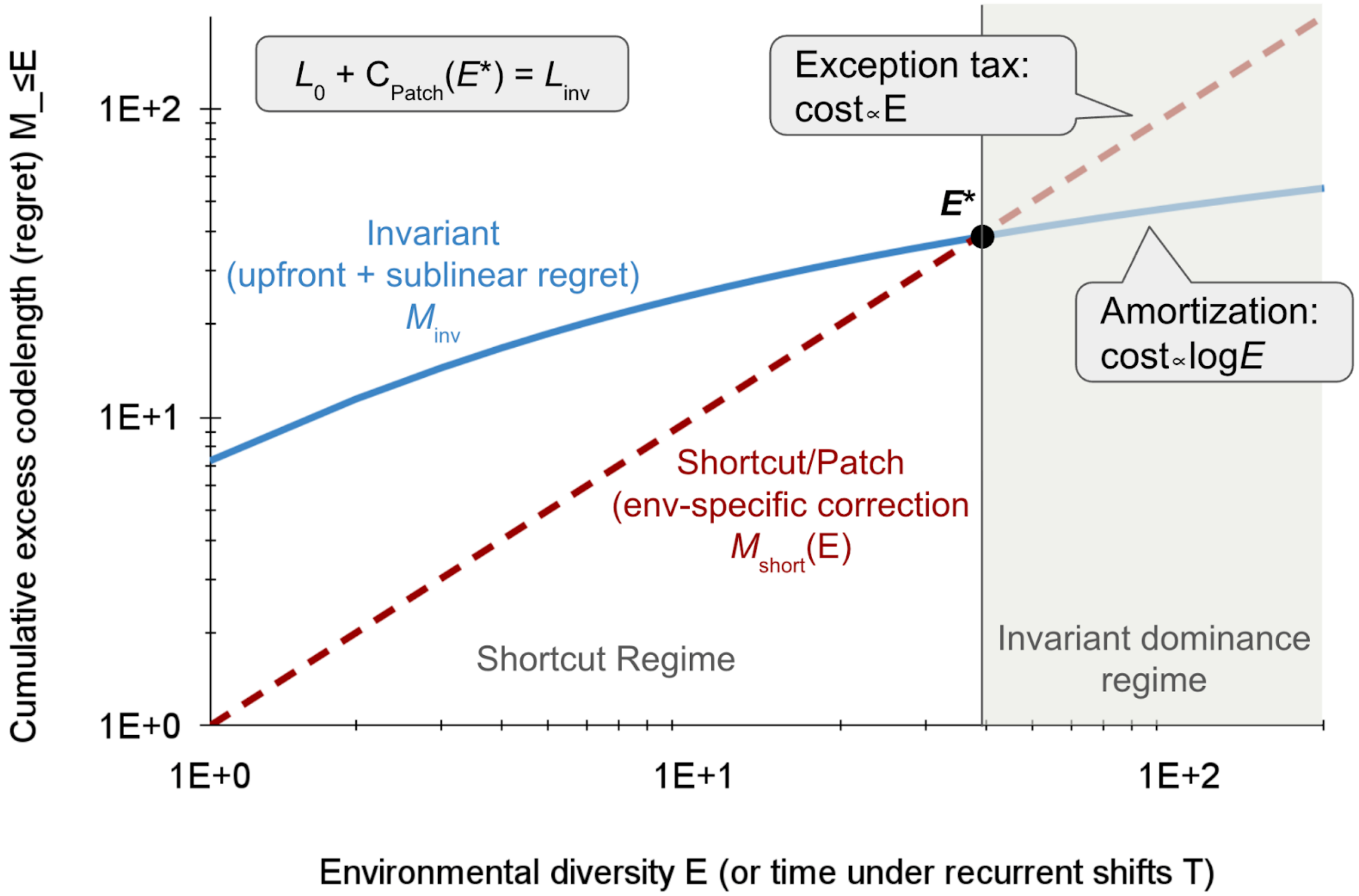

**Figure 2. Exception tax scaling and crossover E* between shortcut patching and invariant compression (schematic).** Cumulative excess codelength ΔL(ε) (regret) is plotted against environmental diversity E (the effective number of distinct contexts induced by the shift family ε; equivalently, time T under recurrent shifts) on log–log axes. A shortcut/patching strategy M_short (red, dashed) has low initial cost but accumulates an exception tax that grows approximately linearly with diversity (ΔL(ε) ∝ E), reflecting repeated patching across new contexts. An invariant strategy M_inv (blue, solid) incurs higher upfront structural cost but amortizes updates across shifts, yielding sublinear growth (illustrated schematically as ΔL(ε) ∝ log E). The intersection defines the crossover threshold E* (inset: $L_0$ + C_patch(E*) = L_inv), beyond which invariants dominate as the compression-efficient solution (shaded region).

# 3. Biological evidence: ITI and CEP in neural substrates

Section 2 framed ITI as a persistence-driven constraint toward predictive compression, and CEP as the shift-based mechanism that selects shift-stable invariants (and, under intervention-rich shifts with modular mechanisms, causal-generative structure). This section reviews biological evidence consistent with that channel—not as metaphor, but as measurable signatures of frontier-seeking under resource limits and of representational strategies shaped by environmental shift.

## 3.1 Energetic constraint: why prediction and representation are expensive

The human brain consumes approximately ~20% of resting metabolic energy despite comprising ~2% of body mass [11]. For ITI, the relevant point is that neural tissue is costly because high-bandwidth signaling and synaptic computation require continuous maintenance of ion gradients, transmitter cycling, and membrane charging. Attwell & Laughlin's energy budget for grey matter identifies action potentials, synaptic currents, and presynaptic release as dominant drivers of ATP demand [31]. In other words, a large fraction of energy is spent on information-bearing activity.

ITI does not require that metabolic cost map linearly to mutual information. The biologically defensible claim is weaker and sharper: metabolic limits impose a steep cost gradient on representational and update strategies, penalizing redundant activity and non-predictive degrees of freedom. Classic analyses emphasize that energy constraints force sparse, high-value signaling rather than dense codes [31,32]. Under ITI, this "bit budget" makes predictive compression consequential rather than optional.

CEP adds the shift dimension: in nonstationary environments, representational strategies that rely on shift-unstable correlations tend to incur persistent mismatch and repeated reconfiguration. Biologically, this "exception tax" is paid not only in storage but in repeated signaling and plasticity (synaptic update processes with real energetic costs) [13]. Thus, selection pressure should favor representational strategies that reduce mismatch per unit metabolic expenditure, especially under recurring distribution shift.

**Epistemic status of the metabolic–codelength link.** We do not claim a derived, quantitative mapping from ATP expenditure to bits of excess codelength. What we claim is weaker and, we believe, sufficient: (i) that metabolic costs of signaling and plasticity are monotonically related to representational and update complexity in a way that makes inefficient strategies consequential, and (ii) that this monotonic relationship is enough to create directional selection pressure favoring shift-stable representations. The quantitative commensurability of metabolic units and information-theoretic units remains an open empirical question (Section 7.5). Our predictions (Section 6) are therefore formulated in terms of scaling behavior (linear vs sublinear growth of integrated mismatch/update proxies) rather than absolute bit-rate conversion.

## 3.2 Frontier-seeking in sensory coding: coding efficiency under naturalistic stimuli

Biology provides unusually direct evidence that early sensory systems operate near information-theoretic bounds [10,49], consistent with ITI's claim that strong resource constraints push systems toward frontier-like trade-offs.

A canonical example is fly photoreceptors, where response nonlinearities appear matched to the statistics of natural luminance contrasts in a way that is close to optimal for information transmission under measured noise constraints [10]. A second, more explicitly information-theoretic case is the blowfly motion pathway: Bialek and colleagues analyzed motion-sensitive neurons under naturalistic stimulation and quantified high coding efficiency [49]—large information transmission rates relative to the entropy rate of the spike train and to theoretical bounds given measured noise—reported as a substantial fraction of the relevant bound, depending on stimulus ensemble and estimator details [49]. Here "efficiency" is best read as how effectively the spike train uses its entropy budget to carry stimulus-relevant information, not as a generic Shannon channel-capacity claim.

These results align with ITI in two ways:

1. **Tight bottlenecks.** Early sensory stages and small nervous systems face hard metabolic and wiring constraints; wasted spikes are fitness-costly [10].
2. **Task-directed coding.** Optimization is not for reconstructing raw input per se, but for transmitting behaviorally relevant structure (e.g., motion signals relevant for control) [13].

CEP adds a mechanistic reason such efficiency should privilege certain features: as an organism moves, sensory statistics are continuously perturbed by shift families (illumination, viewpoint, self-motion, occlusion). Codes that depend on shift-unstable correlations should generate sustained residual uncertainty and update demand as contexts vary; codes aligned to shift-stable structure should amortize that cost across contexts.

Sparse coding work provides a complementary demonstration of the same pressure at the representational level: learning sparse codes on natural images yields V1-like receptive fields, suggesting that efficiency constraints on naturalistic data can recover canonical early-vision structure [50].

## Table 3.1. Biological evidence and alternative explanations (and how ITI/CEP distinguishes them)

**Purpose:** PLR readers will reasonably ask whether the cited biological phenomena require ITI/CEP or are merely consistent with it. This table summarizes key alternatives and the specific empirical signatures ITI/CEP uses to discriminate.

| Phenomenon / evidence | ITI/CEP interpretation | Alternative explanations (representative) | CEP-distinguishing test (this paper) |
|---|---|---|---|
| High coding efficiency in early sensory pathways (e.g., fly photoreceptors; blowfly motion coding efficiency under naturalistic stimulation) [10,49] | Tight bioenergetic/wiring budgets create strong pressure toward frontier-like trade-offs; under shift, efficient codes preferentially preserve shift-stable predictive structure | Efficient coding as redundancy reduction without a shift-regret mechanism; redundancy/overcompleteness as robustness/error-correction under noise and correlated variability (e.g., population coding/noise-correlation perspectives [51.52]; cf. redundancy-for-robustness arguments) | Under controlled shifts that invalidate shortcuts while preserving invariant structure, compare cumulative mismatch/update proxies across increasing environment diversity (E): shortcut-like strategies show approximately linear growth; invariant strategies sublinear (Pred. 2; App. B cue-flip/MMN) |
| Sparse, V1-like receptive fields from efficient coding of natural images [6] | Efficiency constraints on naturalistic data recover canonical low-level features; under ITI/CEP, these features are favored insofar as they are stable under natural transformation families (ε) | Overcomplete or redundant codes can be optimal for robustness and downstream readout; sparse coding not uniquely predicted in all regimes | Shift-family dependence: change the experienced (ε) (e.g., transformation statistics during training), and predicted invariances/features should change accordingly (Pred. 3) |
| Ventral stream hierarchy yields increasing tolerance/selectivity [1,12] | Hierarchy implements progressive extraction of invariants under a natural (ε); active sensing supplies intervention-like diversity that strengthens mechanism alignment under modularity assumptions | Task/data sufficiency explanations: similar invariances can arise from exposure to natural images and task training without invoking a general constraint channel | Restrict or alter active sampling / transformation exposure and measure impact on tolerance and cumulative mismatch/update costs; ITI/CEP predicts measurable degradation when intervention-like diversity is reduced (Pred. 4) |
| Large brain energy use; steep marginal costs for signaling and updating [13,31,32] | ITI/CEP does not explain the entire resting budget; it predicts that steep marginal costs of signaling/plasticity impose selection pressure against strategies requiring broad, repeated patching under shift | Significant fractions of energy support homeostasis/maintenance; redundancy may be tolerated/selected for robustness; representational "expansions" under expertise can occur | Test marginal cost predictions: during controlled shift adaptation, integrated mismatch/update proxies should track the predicted scaling separation, independent of baseline maintenance costs (Pred. 2; App. B controls for salience/complexity) |

### 3.3 Hierarchical tolerance as shift-stable compression - and why “active” perception matters

The ventral stream exhibits a progression from early feature extraction to tolerant object representations—“tolerance” here denoting robustness of neural responses to nuisance transformations such as position, scale, and viewpoint [12]. Importantly, this progression is not merely “compression”; it is selective compression under a natural shift family induced by movement and changing viewpoint. In our framework, such tolerance is a candidate mechanism for achieving conditional-distribution invariance ($P_e(Y|Z)$ stable across ε): if neural responses remain predictively informative despite nuisance variation, the representation approximates shift-stability in the CEP sense.Empirically, selectivity and tolerance both increase as representations propagate from V4 to IT [12], consistent with the idea that downstream representations preserve task-relevant information while discarding nuisance variation [1].

CEP reframes this as a shift problem. Let ε denote the family of transformations an organism experiences (position, scale, viewpoint, lighting, background, partial occlusion). Representations tied to pixel-level correlates of category are shift-unstable under ε; object-centric representations are comparatively shift-stable. CEP predicts that shift-stable representations amortize description length across contexts and reduce long-run regret-like costs (repeated patching).

Critically, biological perception is not passive sampling from a fixed P(X,Y). Active sensing (saccades, head movements, locomotion, manipulation, whisking) expands observational coverage by generating diverse views that decorrelate nuisance factors from task-relevant structure [53]; for example, viewpoint changes can dramatically alter retinal position and local texture statistics while preserving object identity and many object-level regularities. This decorrelation pressure—the systematic variation of nuisance factors while stable structure remains consistent—amplifies the selection for shift-stable representations by enriching ε. Active sensing is not a necessary condition for CEP to operate (Section 4 shows that engineered shift diversity achieves analogous effects), but it is a natural and powerful mechanism by which biological systems generate the shift diversity that makes CEP's selection pressure bite.

When such active exploration approximates the conditions for the causality bridge (intervention-rich shifts + modular structure, Section 2.3), the selected invariants may correspond to causal mechanisms. The degree to which biological active sensing provides this type of shift diversity can be quantified using the protocols in Section 6.

### 3.4 Predictive coding as a candidate implementation: from instantaneous errors to cumulative "regret proxies"

Predictive coding proposes that cortical hierarchies exchange predictions and propagate residual errors upward [7,42,54]. ITI/CEP does not require predictive coding as the unique implementation, but predictive coding provides a biologically plausible way to implement resource-bounded prediction by transmitting what is surprising rather than what is predicted.

To align with Section 2's formalism, we distinguish instantaneous vs cumulative quantities:

- **Instantaneous prediction error** corresponds to a per-sample residual surprisal / unpredicted component—an instantaneous proxy for "uncompressed bits."
- **Regret/excess codelength is cumulative.** The biologically relevant proxy is therefore the time-integral of error-related activity (or the cumulative metabolic cost of error-driven updates) over exposure to a shift family.

On this view, CEP's scaling claim translates into a biological signature: under recurring distribution shifts, correlational/shortcut representations should yield persistently elevated integrated error/update cost (linear in contextual novelty), whereas shift-stable representations should yield integrated residuals dominated by estimation/irreducible uncertainty (sublinear scaling). Empirical measures such as mismatch responses, adaptation dynamics, and repetition suppression are candidate observables for these residual/update costs [55]; Section 6 specifies concrete protocols and falsifiers (including MMN integration windows and cue-flip paradigms).

### 3.5 Biological summary: how these signatures bear on the convergence puzzle

The biological evidence reviewed here supports three core links in the ITI→CEP channel:

1. Resource constraints are real and steep in neural tissue [31,32], supporting ITI's premise that representational efficiency is consequential.
2. Frontier-seeking is empirically visible in sensory coding as high coding efficiency under naturalistic conditions [10, 49], consistent with strong selection against wasted signaling.
3. Hierarchical tolerance in ventral vision can be interpreted as shift-stable compression under a natural transformation family (ε): increasing tolerance to nuisance transformations along V4→IT [1,12] is consistent with progressive achievement of conditional-distribution stability (Section 1.4), with active sensing supplying intervention-like diversity that strengthens the CEP → mechanism alignment claim.

4. These observations do not by themselves prove the conditional causality bridge detailed in Section 2.3, which requires intervention-rich shifts and modular causal structure. But biological perception is intrinsically active and nonstationary, making brains a particularly stringent testbed for CEP: the rich shift family generated by active sensing should produce strong and measurable separation between shortcut and invariant strategies (Section 6). Conversely, artificial systems trained with limited shift diversity provide a natural control—CEP predicts weaker invariant selection and greater shortcut vulnerability in such regimes (Section 4.4).

# 4. Artificial systems: compression efficiency under engineered objectives and shifts

Section 3 argued that biological neural systems operate under steep energetic gradients that make predictive compression a viability constraint (ITI) and that active sensing supplies intervention-like diversity that can make shift-stable structure mechanism-aligned (CEP with ICM-rich shifts). Artificial neural networks differ in substrate and selection pressures, but they provide an unusually tractable testbed: we can vary objectives, architectures, data diversity, and shift families, and measure how representations change. This section reviews evidence that modern deep networks exhibit the same qualitative separation emphasized by CEP: correlational/shortcut strategies succeed in-distribution but pay sustained excess codelength under shift, whereas strategies trained to encode shift-stable structure reduce this shift regret.

### 4.1 Predictive loss as codelength: the simplest cross-substrate bridge (with a formal qualifier)

Most modern deep learning is trained by minimizing negative log-likelihood (cross-entropy) of a normalized predictive distribution. Under this assumption—i.e., when the loss is $\ell_t = -\log p_\theta(y_t|z_t)$—the objective is directly interpretable as predictive codelength (up to the choice of log base), and cumulative loss corresponds to cumulative codelength [20,21]. This mapping is not universal (e.g., hinge loss or MSE without an explicit likelihood model), but it covers the dominant regimes for classification and language modeling that drive the current convergence puzzle.

Two consequences follow:

1. **Training is compression of the training distribution.** Improving held-out negative log-likelihood is improving compression of unseen samples from the same (or closely related) distribution [20,21,56,70,71].
2. **Distribution shift is excess codelength.** When evaluation moves from P to Q, the increase in expected log-loss is precisely an excess codelength term. This is the in-silico analogue of CEP's "exception tax."

This allows CEP to be tested without high-dimensional mutual information estimation: the central empirical object is shift-regret, operationalized as the increase in predictive codelength across a shift family ($\varepsilon$).

### 4.2 Frontiers and scaling: what we claim (and what we do not)

Deep learning exhibits reproducible empirical regularities in which predictive codelength decreases as model capacity, data, and compute increase within major architecture families. For PLR purposes we state a conservative, citation-stable claim:

For major model families in vision and language, there exist reproducible performance–resource frontiers: increasing model capacity and compute can systematically reduce held-out predictive loss and improve performance over wide ranges, and improvements are often well-approximated by simple functional forms, as demonstrated across contemporary architectures [5,53,54].

This conservative statement is supported by peer-reviewed large-scale studies in language and vision, including demonstrations that predictive loss and transfer performance improve predictably with scale in contemporary architectures [8].

**What we do not claim.** ITI/CEP does not require that deep networks literally "compress I(X;Z)" during training, nor that an Information Bottleneck narrative accurately describes all deep learning regimes. In particular, work including Saxe et al. [44] shows that in many common training settings (e.g., deterministic ReLU networks without explicit bottlenecks), the empirical claim "I(X;Z) compresses during training" is not generally supported and may hinge on estimation artifacts or architectural choices.

**How ITI/CEP uses the frontier idea.** CEP uses frontier language in an operational sense: for fixed task families and resource constraints, some representational strategies achieve lower predictive codelength for the same resource cost. The relevant efficiency measure is therefore frontier-relative, not absolute: how much predictive codelength is reduced per unit representational budget (parameters, compute, memory), evaluated within the same task/shift family.

This reframes the IB controversy as supportive rather than threatening: if resources are abundant, systems can afford "patches" (context-specific memorization) without immediate penalty; CEP predicts that the pressure to encode shift-stable invariants becomes dominant when constraints bind (limited capacity, costly adaptation, or strong shift diversity).

## 4.3 Engineered shift families: data augmentation and self-supervision as proxies for active sensing

A direct way to test CEP is to engineer ε: apply transformations that simulate nuisance variation (crops, rescaling, color shifts, blur, occlusion) and evaluate whether learned representations become invariant to those perturbations. In vision, both augmentation-heavy supervised training and self-supervised contrastive methods can be interpreted as enforcing invariance across a prescribed shift family [59,60].

A central ITI/CEP point is that **data augmentation is an engineered proxy for active sensing**. Movement and saccades generate shifts in biological input statistics; augmentation generates analogous perturbations in silico. If CEP is correct, both systems—trained under comparable shift families—should be channeled toward similar invariances. This provides a principled route from "augmentation improves robustness" to the more specific convergence claim: augmentation induces the same kind of nuisance-factor variability that biological agents generate through action.

CEP also predicts which invariances emerge: invariance should track the experienced shift family, not a universal object prior. This shift-family dependence is falsifiable: change ε, and the learned invariances should change accordingly.

### 4.4 Shortcut learning: correlational compression with high shift-regret

Modern deep networks often exploit "shortcuts": features that are highly predictive in the training distribution but do not reflect shift-stable structure and fail under distribution shift. This is a direct instantiation of CEP's shift-unstable correlation class. The shortcut learning literature documents that standard empirical risk minimization can lock onto spurious correlates (e.g., texture cues, background regularities), yielding strong in-distribution performance while degrading under OOD evaluation [9].

OOD and robustness benchmarks provide operational proxies for shift-regret. For example:

- **ImageNet-C** quantifies robustness to common corruptions using standardized corruption error metrics [9]
- **ObjectNet** and related dataset-shift benchmarks test generalization when image composition and object presentation differ from ImageNet [63]
- **WILDS** formalizes real-world distribution shifts across multiple domains with standardized evaluation protocols [64].

In CEP terms, these suites approximate exposure to new $e \in \varepsilon$ and measure the excess codelength incurred when correlational compression is transported outside its training regime. Repair typically requires additional "patching" mechanisms—more data diversity, stronger augmentations, explicit invariance objectives, or fine-tuning—precisely the kinds of context-specific corrections CEP predicts will be required when the representation is not shift-stable.

### 4.5 "Grokking" as an idealized illustration of CEP's mechanism (explicitly suggestive, not load-bearing)

A suggestive phenomenon in artificial training dynamics is "grokking," where networks first fit training data with poor generalization and later transition—sometimes abruptly—to strong generalization without architectural change [69]. Grokking has been demonstrated most clearly in small algorithmic tasks and is not known to be universal in natural-data regimes. For this reason, we treat it as an idealized illustration rather than core evidence.

Nevertheless, grokking is conceptually aligned with CEP's separation between (i) correlational/patch solutions that succeed in-distribution and (ii) invariant/rule-like solutions that reduce shift-regret. In the language of Section 2, grokking resembles a transition from effectively linear regret under a controlled shift family to sublinear regret once the model internalizes structure that transfers. We revisit this as a test protocol (not as an empirical cornerstone) in Section 6, where we can define shifts that explicitly flip shortcuts and measure regret scaling.

## 4.6 In-context learning: compression-as-inference (scope-limited)

Large language models exhibit in-context learning: given a prompt containing examples or instructions, the model can adapt behavior without parameter updates [63]. A conservative interpretation compatible with ITI/CEP is amortized inference: the model compresses the prompt into a latent structure that predicts the continuation distribution.

We do not require a specific mechanistic theory of in-context learning here (and several are actively debated). For ITI/CEP, the operational claim is narrower: in-context learning can reduce shift-regret by rapidly selecting or constructing the appropriate conditional predictor when the "environment" changes within a sequence. This is testable via controlled synthetic shifts where shortcut cues are flipped and the correct invariant changes.

## 4.7 AI summary: what artificial systems add to mutual validation

Artificial systems strengthen the ITI→CEP argument in three ways:

1. **Substrate-independence under controlled manipulation.** We can vary compute, data, and shift families and observe systematic changes in OOD performance and shift-regret, testing CEP directly [57,59].
2. **Direct measurement of excess codelength.** When models are trained with normalized likelihood objectives, predictive loss is codelength [20,21]; "exception tax" is directly observed as excess loss under shift and the extra complexity required to repair it.
3. **A complementary failure regime.** Unlike biological systems, many AI systems are trained largely passively and without intervention-rich diversity; CEP predicts increased shortcut vulnerability in such regimes. Conversely, engineered proxies for active sensing (augmentation, multi-environment training, interactive data generation) should reduce shortcut reliance by enriching ε [9,55].
4. **Reinforcement learning from human feedback (RLHF) as a shift-family narrowing intervention.** Reinforcement learning from human feedback (RLHF) as an evolutionary bottleneck. Just as biological domestication creates evolutionary bottlenecks that favor human-preference traits over general survival traits, RLHF substitutes a preference-congruent target for the broad predictive codelength objective [64]. Human preference is shaped by cognitive bias and narrative coherence, which do not constitute an intervention-rich shift family (Section 2.3). The result is a compressed representation that is preference-stable rather than shift-stable: the model selects features invariant under the distribution of human raters, not under the intervention-rich diversity that would favor mechanism-aligned representations. CEP predicts that this narrowing acts like domestication—RLHF-trained models should exhibit systematically elevated shift-regret on out-of-distribution tasks relative to models trained under unmodified predictive loss.

**Human Engineering as a Simulator of Physical Selection.**

In artificial systems, gradient-based optimization is the inner-loop parameter update. The convergence on primate-like representations is not evidence of autonomous AI evolution; rather, it demonstrates the substrate-independence of the ITI constraint. The outer-loop selection mechanism operates through deployment: models that fail to predict adequately under real-world distribution shifts are replaced by competitors that track reality more faithfully — a selection dynamic analogous to biological fitness. When training pipelines incorporate diverse, multi-environment regimes (aggressive data augmentation, distribution shifts) under resource budgets $R(m)$, they compress the timescale of this selection by exposing models to shift diversity before deployment. Configurations with high shift penalties — those requiring many context-specific patches — are outcompeted by those that amortize prediction via shift-stable invariants. The network discovers generative invariances because the same constraint that penalizes biological shortcuts (cumulative codelength under shift) penalizes artificial ones.

In Section 5 we synthesize biological and artificial evidence into a single convergence argument and emphasize informative divergences as differing operating points on a shared resource–performance frontier. Section 6 specifies protocols that isolate CEP's predictions in controlled shift families, with explicit falsifiers.

# 5. Synthesis: the ITI → CEP constraint channel and why convergence is evidence

Sections 1–4 established (i) the convergence problem (quantitative brain–model alignment plus broad representational parallels), (ii) a theoretical mechanism for why efficient prediction under shift selects invariants (regret/excess codelength), and (iii) evidence that biological and artificial systems instantiate the same trade-offs under different resource constraints. This section synthesizes these elements into a single constraint channel that resolves the central mystery: why brains and deep networks converge.

## 5.1 The grand causal chain (constraint channel, not teleology)

We summarize the argument as a chain of constraints. No link requires appeal to intention or design; each link follows from physical, information-theoretic, or selection constraints.

**[Persistence ⇒ Prediction ⇒ Compression ⇒ Efficiency under budgets ⇒ Shift-stable invariants]**

Under the additional conditions detailed in Section 2.3—intervention-rich shift families and approximately modular causal structure—these shift-stable invariants tend to correspond to causal mechanisms. Without those conditions, CEP still selects shift-stable representations, but these need not be causal (Section 2.3.1).

**Link 1: Persistence → Prediction.**

Active persisters must maintain viable states against perturbations. In uncertain environments, this characteristically requires anticipatory regulation—i.e., prediction—because reactive control without predictive structure cannot reliably keep essential variables within bounds under the resource and uncertainty conditions typical of biological and deployed artificial systems [7].

**Link 2: Prediction → Compression.**

Prediction under finite memory/time/energy cannot be achieved by storing growing sensory history. A scalable strategy is to exploit regularities to build compact internal structure—compression in the information-theoretic sense [33,35].

**Link 3: Compression → Efficiency (under budgets).**

Compression must be achieved under resource budgets. In biological substrates, signaling and synaptic computation carry steep metabolic costs [11,13,31,32], which penalize redundant or non-predictive degrees of freedom—though the quantitative mapping from metabolic expenditure to information-theoretic cost remains an open problem (Section 3.1). The directional prediction—that metabolically costlier strategies face stronger selection pressure—is sufficient for ITI/CEP’s constraint logic without requiring exact commensurability. In artificial substrates, compute/memory budgets impose analogous constraints and create observable performance–resource frontiers (Section 4.2) [8,57,58]. This stage is where ITI becomes concrete: not every predictive strategy is viable—only those whose predictive gains justify their representational and update costs.

**Link 4: Efficiency under shift → invariants (CEP).**

In stationary settings, many compressions can look good. Under distribution shift, representational strategies diverge. CEP's key formal claim is a separation in excess codelength (regret) scaling: shortcut strategies that exploit shift-unstable correlations incur approximately linear regret in environmental diversity, whereas invariant representations incur sublinear regret dominated by estimation/approximation rather than patch accumulation (Section 2.2). This is the "exception tax" mechanism in formal form.

**Conditional extension: Invariants → causal mechanisms (under specific environmental conditions).**When the environment provides intervention-rich shifts (perturbing different factors independently) and the underlying causal structure is approximately modular (mechanisms can be changed independently), the shift-stable invariants selected by CEP tend to align with causal mechanisms—in the operational sense that they support intervention-invariant prediction and localized updating [28,29].

This is because modular mechanisms are precisely the representational units that remain predictively valid across interventional shifts while requiring only localized updates (Section 2.3). This extension rests on empirical premises about environmental structure rather than information-theoretic necessity.

## 5.2 Why the synthesis is necessary (not merely sufficient)

A recurring objection is that each link could be true independently without explaining convergence. The ITI→CEP synthesis matters because it closes two gaps that remain open in neighboring frameworks.

**ITI without CEP is incomplete.**

ITI alone explains why predictive compression is pressured, but not what compression finds under shift. A system could be predictive in-distribution via correlational strategies that fail catastrophically out of distribution. ITI does not by itself explain why robust representations should emerge rather than patchwork heuristics.

**CEP without ITI lacks force.**

CEP alone is a statement about which representations minimize regret under shift. But why should any system care about this? ITI supplies the selection pressure: persistent systems (biological by survival selection; engineered by deployment and evaluation) cannot sustain strategies that pay unbounded "exception taxes" under recurring shift.

Together, they explain convergence: **the set of viable solutions is narrowed by physical constraints (ITI), and within that set the stable solutions are those that minimize regret across shifts (CEP).**

## 5.3 Resolution of the three biological mysteries

Section 1 opened with three biological puzzles that demand a unified physical explanation. The ITI→CEP channel resolves each.

**Mystery 1: Why the brain's metabolic budget is so large (~20% resting energy)**

**Observation.** The brain is metabolically expensive, but its resting budget includes substantial baseline maintenance and homeostatic costs in addition to signaling [13,31].

**ITI→CEP resolution (qualified).** ITI/CEP does not claim to explain the entire ~20% allocation. Rather, it uses the empirically steep marginal cost of signaling and updating—spikes, synaptic currents, and plasticity—to argue that inefficient representational strategies (those requiring persistent mismatch and broad reconfiguration under shift) are selectively penalized. CEP predicts that, under recurring shift, representations that rely on unstable correlations impose sustained "exception taxes" in integrated mismatch/update cost, whereas shift-stable representations amortize these costs.

**Dual-component "exception tax."** Under CEP, the cost of relying on shift-unstable correlations can be paid through two distinct physical channels: (i) an ongoing energy-per-prediction cost from residual mismatch signaling [31] and (ii) an episodic energy-per-update cost from plasticity and reconfiguration required to incorporate patches or revise representations. Shift-stable representations reduce both components by lowering residual error rates and localizing the updates required when environments change (Appendix B; Section 6 cue-flip/MMN and adaptation protocols).

Importantly, the prediction is not that "smarter brains burn less energy in total," but that they achieve lower residual uncertainty and lower update burden per unit predictive performance—made operational in Section 6 via integrated error/update cost proxies and crossover concepts.

**Mystery 2: Why insect sensory coding can approach theoretical limits**

**Observation.** Sensory pathways can exhibit high coding efficiency under naturalistic stimulation, with classic cases in fly photoreceptors and motion-sensitive neurons [10,49].

**ITI→CEP resolution.** Under severe resource constraints (small nervous systems, tight wiring, high noise), the efficiency frontier is sharply binding. Systems that waste spikes on redundant or non-predictive bits pay immediate fitness costs. ITI explains why selection pressure is steep; CEP explains what kind of structure is worth encoding under shift: invariants that amortize prediction across the organism's natural transformation family. The fly's near-efficient code is therefore not a curiosity; it is an expected consequence of strong constraint acting on a predictive channel.

**Mystery 3: Why the ventral stream builds a hierarchy that yields invariance**

**Observation.** Along the ventral stream, selectivity and tolerance increase, yielding representations that are robust to nuisance transformations [1,12].

**ITI→CEP resolution.** The organism experiences a shift family (ε) induced by movement and saccades. A representation tied to pixel-level correlates is shift-unstable and would pay persistent regret as viewpoint, illumination, and background change; object-centric representations are shift-stable and amortize description length across ε. Hierarchy is a plausible strategy for extracting invariants progressively under resource constraints: each stage factors a subset of nuisance variability, reducing the mismatch burden passed forward.Active sensing enriches the shift family ε by systematically decorrelating nuisance factors (viewpoint, illumination, background) from stable object properties, strengthening the selection pressure for shift-stable representations [53]. When the resulting shift family approximates the conditions specified in Section 2.3 (intervention-rich diversity + approximate mechanism modularity), these representations may additionally align with causal mechanisms—but the primary CEP prediction (shift-stability, not causality) holds regardless.

## 5.4 Why convergence is evidence (mutual validation logic)

The convergence of biological and artificial systems is not just a curiosity; it functions as a mutual validation test of the claim that the constraint is physical.

- **Biology demonstrates reachability under selection.** Over evolutionary time, systems subject to metabolic and ecological constraints converge toward efficient predictive representations.
- **AI demonstrates substrate-independence.** With different substrates and optimization dynamics, systems optimizing predictive codelength under constraints rediscover similar representational motifs and, in key cases, align quantitatively with neural responses (Section 1).

The combination is stronger than either alone. If the convergence were a biological accident, engineered systems would not reliably rediscover it. If it were an engineering artifact, evolution would not have repeatedly found it under different constraints. The most parsimonious explanation is that both are being constrained by a common channel: **efficiency of prediction under finite resources and distribution shift**. Here "shared frontier" is not a claim that ATP and FLOPs are directly commensurable; it is a claim about the shared topology and scaling of trade-offs (excess codelength under shift, crossover structure as ε becomes richer) across substrates.

## 5.5 Informative divergences: different operating points on the same frontier

Convergence does not imply identity. ITI/CEP predicts both similarities and divergences because different substrates impose different budgets and therefore different operating points on a shared frontier.

**Metabolic sparsity vs compute abundance.** Brains face tight energy/wiring constraints and thus emphasize sparse signaling and local update rules [31,32]. Many artificial systems can afford dense overparameterization during training. CEP predicts that shortcut patching is more tolerable when resource costs are low; invariants become dominant when constraints bind or shift families are rich.

A related and potentially confusing phenomenon in machine learning is **overparameterization** (including "double descent"), where increasing model capacity can improve generalization rather than simply enabling patch accumulation. In ITI/CEP terms, adding parameters can have two qualitatively different effects: it can (i) lower the effective patch cost $c_{patch}$ in a digital substrate (pushing $E^*$ outward, making patching cheap for longer), and/or (ii) enlarge the hypothesis class so that invariant structure becomes representable and easier to discover—effectively moving the system from a shortcut-dominated regime toward an invariant-capable regime. Thus, capacity growth improving OOD performance does not contradict CEP; the distinguishing signature remains how excess codelength scales with environmental diversity and whether improvements persist under shortcut-flipping shifts at fixed evaluation protocols [65]. In the extreme limit where additional capacity is effectively free (very low marginal $c_{patch}$), the crossover $E^*$ can be pushed beyond feasible training exposure, allowing shortcut-heavy strategies to persist much longer than in metabolically constrained biological systems.

**Active vs passive data generation.** Biological agents generate intervention-like shifts through action; many AI systems train passively on fixed corpora. CEP predicts greater shortcut vulnerability in passive regimes and improvement when training incorporates richer shift families or interactive data collection (Sections 3.3–4.4).

These divergences are not embarrassments; they are confirmations of a constraint-based account. They are also predictive: altering budgets or shift richness should shift systems along the same trade-off curve.

## 5.6 Synthesis as a testable program

The ITI→CEP channel is a compact theory only if it earns its keep empirically. It predicts:

- **Efficiency/shift-regret → OOD robustness coupling** within task families (across biological and artificial systems).
- **Regret scaling separation** under shortcut-flipping shifts: approximately linear vs sublinear regimes (Section 2.2.3; operationalized in Section 6).
- **Shift-family dependence of invariances:** invariances track the experienced (ε), not a universal prior.
- **Action-generated diversity strengthens mechanism alignment:** active sensing/interactive data improves invariant/mechanism learning relative to passive sampling, all else equal.

Section 6 turns these into protocols with explicit falsifiers.

# 6. Testable predictions and protocols: substrate-symmetric tests of ITI→CEP

Sections 1–5 motivate ITI→CEP as a constraint channel explaining why biological and artificial neural systems converge on similar representational strategies. This section specifies quantitative, falsifiable predictions and core empirical tests. We state each prediction using the manuscript's formal objects—**predictive codelength** and **excess codelength (regret)** under a shift family (ε)—and we make the biological measurements concrete enough to be actionable.

## 6.0 Measurement principle: prefer codelength and shift-regret over high-dimensional MI

High-dimensional mutual information is difficult to estimate robustly and can be sensitive to estimator choices [45,46,47]. By contrast, in the dominant modern training regimes where the loss is a negative log-likelihood of a normalized predictive distribution, $\ell_t = -\log p_\theta(y_t|z_t)$, cumulative loss is cumulative predictive codelength [20,21]. Therefore, the primary operational quantities are:

- **Baseline predictive codelength:** $L_P(m) = E_{(x,y)\sim P}[-\log p_m(y|x)]$.
- **Shift penalty / excess codelength:** $\Delta_e(m) = L_{Q_e}(m) - L_P(m)$, where $Q_e$ is a shifted distribution.
- **Cumulative shift-regret:** cumulative excess codelength across a sequence of shifts, defined with an explicit reference class (next subsection).

**Note on resources across substrates (what "shared frontier" means).** We do not assume ATP/Joules, parameters, and FLOPs are directly commensurable. Predictions in Section 6 are tested within a substrate using a chosen resource proxy $R(m)$. When we speak of a "shared frontier," we mean shared topology of the trade-off (frontier shape, scaling, and crossover structure under increasing shift richness), not equality of raw resource units. Cross-substrate comparisons should therefore use frontier-normalized, dimensionless quantities and scaling variables such as $E/E^*$, rather than comparing ATP to FLOPs directly.

Mutual-information-based quantities remain useful as conceptual scaffolding (Section 2.4.3), but none of the core falsifiers or protocols in this manuscript depend on mutual information estimation.

## 6.1 Regret definitions: reference classes and what "linear vs sublinear" means here (critical)

Let M denote a model class (representation + predictor family), and let the system experience environments $e_1,\dots,e_T$ with samples $(x_t,y_t)\sim P_{e_t}$. Define within-class online regret [37]:

$$\mathrm{Regret}_T(M) \equiv \sum_{t=1}^{T} -\log p_{\hat{m}_{t-1}}(y_t|x_t) - \min_{m\in M} \sum_{t=1}^{T} -\log p_m(y_t|x_t).$$

CEP's scaling separation is fundamentally a comparison between classes:

- **M_inv:** a class capable of representing the invariant/mechanism (shift-stable) structure.
- **M_short:** a class that relies on (or is induced to rely on) shift-unstable correlates and cannot represent the invariant mechanism without patching.

A convenient "between-class" object is the excess codelength gap:

$$G_T \equiv \min_{m\in M_{short}} \sum_{t=1}^{T} \ell_t(m) - \min_{m\in M_{inv}} \sum_{t=1}^{T} \ell_t(m).$$

CEP predicts that under shortcut-flipping shift families, $G_T$ grows approximately linearly with exposure to environmental diversity, whereas $\mathrm{Regret}_T(M_{inv})$ remains sublinear (estimation/approximation dominated). This resolves the "trivial linear regret" concern: linearity is not an artifact of picking a too-weak reference class; it is the signature of a representational strategy that cannot amortize across shifts.

## 6.2 Prediction 1: frontier-relative efficiency predicts OOD robustness (within task families)

**Claim.** Within a fixed task family and shift family (ε), systems that achieve higher frontier-relative efficiency exhibit smaller OOD shift penalties ($\Delta_\varepsilon$).

**Operational definition (task family).** Here we operationalize a "task family" as a fixed input domain and target specification (e.g., ImageNet-1k image classification; a fixed WILDS task) while varying architectures and training regimes within that task. Cross-task comparisons (different domains/targets) should be done only via frontier-normalized quantities (Section 6.7), not by comparing raw resource units.

For a model m,

$$\Delta_\varepsilon(m) = E_{e\sim\mu}[L_{Q_e}(m) - L_P(m)].$$

Let R(m) be a resource measure (parameters, FLOPs, memory, or energy). Define frontier-relative efficiency as a within-family slope, e.g.

$$\hat{\mathrm{Eff}}(m) \equiv -\Delta L_P/\Delta R \text{ or } -\Delta\Delta_\varepsilon/\Delta R.$$

**Note on resources across substrates.** We do not assume ATP, parameters, and FLOPs are directly commensurable. Prediction 1 is tested **within a substrate** (or within a matched experimental system) using a chosen R(m). Cross-substrate comparison, when desired, should use frontier-normalized quantities (distance-to-frontier or Êff_norm) rather than raw resource units.

**Prediction.** Within-task,

**corr(Êff(m), -Δ_ε(m)) > 0.**

**Core empirical tests (AI).** Compare matched-capacity models trained with different shift-family enforcement (standard ERM vs augmentation-invariant or multi-domain training) and evaluate on ImageNet-C [9], ObjectNet [61] and WILDS tasks [62]. Falsifier (example threshold): with adequate power (e.g., $n \geq 50$ trained variants), robust corr<0.2 undermines the efficiency–robustness link.

**Core empirical tests (biology).** Use proxies tied to the CEP mechanism (Section 3.4): cumulative mismatch/update cost during adaptation to shifts.

- **EEG/MEG:** integrated mismatch negativity (MMN) amplitude over repeated shift exposures as a proxy for cumulative residual surprisal [55].
- **fMRI:** repetition suppression / adaptation slopes under systematic shift manipulations.
- **Electrophysiology:** time-integrated spike counts or LFP power in populations implicated in error signaling.

We treat integrated MMN/adaptation measures as candidate proxies for cumulative mismatch cost rather than an established identity. If the predicted scaling separation is observed behaviorally under cue-flip shifts but not in MMN (or not in the chosen neural signal), this would indicate that the regret-like cost is implemented in slower adaptation/plasticity or metabolic processes rather than in fast error-signaling populations, prompting proxy revision rather than immediate rejection of CEP.

**Disconfirmation.** If higher biological "efficiency proxies" (better behavioral performance per unit integrated mismatch/update cost) correlate with worse OOD robustness across controlled shifts, Prediction 1 fails.

## 6.3 Prediction 2 (CEP core): shortcut-flipping shifts produce linear vs sublinear scaling (with a shift-richness condition)

The dominant explanations for why scaling improves reasoning have focused on increased parameter capacity for pattern storage and retrieval, emergent phase transitions from memorization to generalization, or in-context learning as implicit Bayesian inference [8,66,67]. These accounts are descriptive — they characterize that scaling works without explaining why certain capabilities emerge before others or why the improvements are uneven across domains.

A more fundamental mechanism follows directly from MDL principles. Locally-true shortcuts — culturally contingent beliefs, spurious correlations, context-specific heuristics — encode inconsistently across contexts. They cannot be compressed into a single stable representation because they are not stable; each new context requires re-encoding the local variant. Genuinely invariant structure — mathematical truth, physical regularity, logical consistency — encodes identically regardless of context, because cross-context consistency is precisely what makes it invariant. As data volume grows and context diversity increases, the cost differential between these two strategies widens. The compression advantage of invariant structure is not fixed; it compounds with scale. This is why mathematical reasoning emerges robustly early in scaling while social and culturally-embedded reasoning remains shallow and brittle — not because math is simpler in an absolute sense, but because its invariant structure provides an unambiguous, multiply-reinforced compression signal across every domain it touches.

**Shift-richness requirement.** The shift family must be rich enough that patching becomes more expensive than representing the invariant mechanism. A practical criterion is a crossover point $E^*$ where

$$E \cdot c_{patch} > c_{inv},$$

so for $E > E^*$, the invariant strategy becomes strictly better in total codelength.

**Core empirical tests (AI).** Use synthetic multi-environment benchmarks where a shortcut feature flips across environments while the invariant feature remains stable. Compare $M_{short}$ vs $M_{inv}$ explicitly and measure

$$G_E \equiv \min_{m \in M_{short}} \sum_{i=1}^{E} L_{Q_{e_i}}(m) - \min_{m \in M_{inv}} \sum_{i=1}^{E} L_{Q_{e_i}}(m),$$

predicting $G_E$ grows ~linearly in $E$ once $E>E^*$.

**Core empirical tests (biology).** Use a cue-flip paradigm: train on a stable rule (invariant mechanism) and a shortcut cue predictive in $e=0$, then flip shortcut validity across environments ($e=1,...,E$) while preserving the invariant rule. Define biological analogues of codelength gap via behavioral surprise/error and integrated MMN/adaptation cost.

**Disconfirmation.** If shortcut-flipping designs do not yield a robust scaling separation between shortcut-trained and invariant-trained systems—given $E>E^*$ and sufficient samples—CEP's central mechanism is undermined.

## 6.4 Prediction 3: ε-tracking invariances (shift-family dependence)

**Claim.** Learned invariances track the experienced shift family (ε). Change ε, and invariances change in predictable ways. This is consistent with the observation that the ventral stream builds tolerance and selectivity under natural variation [1].

**AI.** Train identical architectures under different augmentation families $\varepsilon_1$, $\varepsilon_2$, $\varepsilon_3$ (background vs geometry vs color). Evaluate invariance metrics and aligned shift penalties ($\Delta_{\varepsilon_k}$).

**Biology.** Manipulate transformation statistics experienced during learning and predict that tolerance properties in downstream representations adjust to the experienced ($\varepsilon$).

**Disconfirmation.** If invariances are insensitive to the experienced ($\varepsilon$) (controlling for confounds), CEP's mechanistic claim is incomplete.

## 6.5 Prediction 4: active/interactive data generation reduces shortcut reliance (richer shift families strengthen invariants)

**Claim.** Systems with access to richer shift families—generated by active sensing in biology, or interactive data generation in AI—will show reduced shortcut reliance and reduced shift-regret compared to passive exposure. Richer shift families systematically decorrelate nuisance factors from stable structure, strengthening CEP's selection pressure for invariants. Under the conditions specified in Section 2.3 (intervention-rich shifts + modular structure), these invariants may correspond to causal mechanisms.

**AI.** Compare passive training on static data to training with controllable nuisance perturbations (movable camera in simulation; environments where the agent varies viewpoint/background). Evaluate shortcut-flipping shift suites and measure $\Delta_\varepsilon$ and $G_E$.

**Biology.** Compare passive viewing vs active exploration (restricted saccades/movement vs free exploration) while controlling exposure; use integrated MMN/adaptation costs and behavioral transfer.

**Disconfirmation.** If active/interactive regimes do not reduce shortcut reliance or shift-regret when confounds are controlled, the interventional-diversity component of the CEP→mechanism bridge is weakened.

## 6.6 Structured vs "causal-poor" environments (scope, to avoid false falsification)

ITI/CEP is a claim about structured environments—those with latent regularities that persist (at least approximately) across a relevant shift family. In "causal-poor" or effectively white-noise environments, no representation can achieve low surprisal beyond the entropy floor; the framework predicts that stable invariants are not discoverable because there are none [33,35].

## 6.7 Optional: frontier-normalized (dimensionless) efficiency for cross-domain comparison

If cross-task or cross-substrate comparisons are desired, a robust approach is to compare distance to the empirical frontier, not raw resource units. Let $L_{frontier}(R)$ denote the lower envelope (best observed codelength) achievable at resource level R within a fixed task and evaluation protocol. Let $L_{base}$ denote a fixed baseline codelength (e.g., class-prior predictor for classification; unigram baseline for language).

Define the dimensionless distance-to-frontier:

$$d_L(m) \equiv [L_P(m) - L_{frontier}(R(m))] / [L_{base} - L_{frontier}(R(m))],$$

and analogously for shift penalties ($\Delta_\varepsilon$) by replacing $L_P$ with $\Delta_\varepsilon$ and using the corresponding empirical frontier. Values near 0 indicate near-frontier performance for the chosen resource proxy; values near 1 indicate far-from-frontier performance.

Under ITI/CEP, convergence claims are primarily about (i) the emergence of similar representational strategies and (ii) scaling structure (e.g., linear vs. sublinear regret regimes and crossover E*), rather than equality of raw resource units across substrates.

## 6.8 Consolidated prediction table (with clarified baselines)

| Prediction | Empirical object | Baseline / reference | Expected signature | Falsified if |
|---|---|---|---|---|
| **P1** Efficiency–robustness coupling | $\Delta_\varepsilon(m)$ | within-task frontier comparisons | higher efficiency → lower $\Delta_\varepsilon$ | robust corr < 0.2 (adequate power) |
| **P2** Regret scaling separation | $G_E$, $Regret_T(M)$ | compare $M_{short}$ vs $M_{inv}$ | $G_E \sim \Omega(E)$; inv class sublinear | no separation for $E>E^*$ |
| **P3** ε-tracking | invariance metrics vs trained ε | altered augmentation/exposure | invariances follow ε | invariances insensitive to ε |
| **P4** Active/interactive advantage | $\Delta_\varepsilon$, $G_E$ reduction | passive vs active/interactive | active reduces shortcut reliance | no improvement controlling confounds |

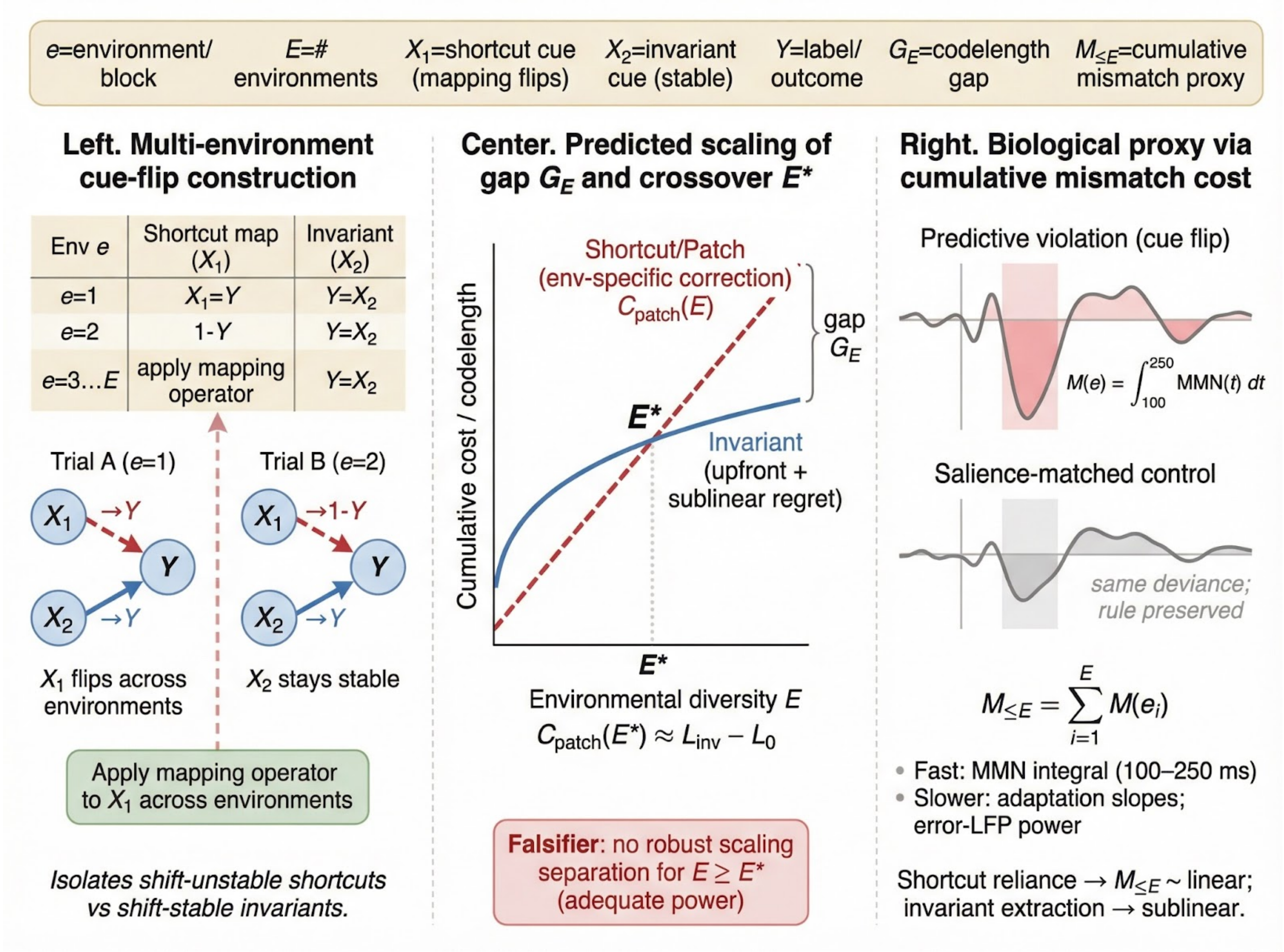

**Figure 3. Protocol logic for the shortcut-flipping paradigm and crossover E*.** Top bar defines notation. (Left) Multi-environment cue-flip construction: the invariant cue $X_2$ maintains a stable mapping to outcome Y across all environments, while a mapping operator flips the shortcut cue $X_1$, isolating shift-unstable correlations from shift-stable invariants. (Center) CEP predicts a scaling separation in cumulative cost: shortcut/patch strategies incur approximately linear growth C_patch(E), whereas invariant strategies yield only sublinear regret, producing a widening codelength gap G_E beyond crossover E* (defined by C_patch(E*) ≈ L_inv − $L_0$). Falsifier: absence of robust scaling separation for E ≥ E* under adequate power. (Right) Biological proxy: cumulative mismatch cost M_{≤E} is constructed from MMN amplitude integrated over a pre-registered window following predictive violations, with a salience-matched control isolating prediction error from sensory deviance. CEP predicts linear M_{≤E} under shortcut reliance, sublinear after invariant extraction.

# 7. Implications and open problems (scientific scope only)

Sections 1–6 develop and operationalize the ITI→CEP constraint channel. This section summarizes implications for neuroscience, AI, and the physics of prediction, and identifies high-leverage open problems.

## 7.1 Neuroscience: explaining neural form as frontier-seeking under shift

**Energy constraints as an explanatory prior (ITI).**

Neuroenergetics provides a steep cost gradient—spikes, synaptic currents, and plasticity are expensive [13,31]. ITI reframes this as a central explanatory prior: energetic costs impose strong constraints [32], suggesting that cortical architectures are shaped toward high predictive performance per unit resource.

**Prediction error as a measurable proxy for "excess codelength."**

CEP's formal object is cumulative excess codelength (regret). In biological systems, the analogue is the time-integrated cost of mismatch and updating (Section 3.4). This suggests a measurement program:

- identify error-related responses under controlled shifts (e.g., MMN [55]),
- integrate these responses across exposure to an increasing number of shift contexts (E),
- test for the predicted scaling separation between shortcut-like and invariant-like regimes (Section 6.3).

**Active sensing as interventional diversity.**

Active sensing generates intervention-like diversity that can make ε ICM-rich in many domains, strengthening mechanism alignment under modularity assumptions [28,29]. Designs that restrict active sampling should reduce invariance acquisition and increase integrated mismatch cost (Prediction 4).

## 7.2 Artificial intelligence: design principles implied by CEP (without claiming universality)

**Prefer objectives that penalize cross-environment regret.**

CEP points to compression efficiency under shift as the relevant engineering lever: training and evaluation that penalize excess codelength across a specified (ε).

**IB controversy as a boundary condition.**

Work including Saxe et al. [44] shows that "I(X;Z) compression during training" is not universal. ITI/CEP predicts that compression-like pressures become dominant when constraints bind and when cross-environment regret is penalized; abundant capacity can mask invariant pressure by allowing patching.

**Shift-family dependence is a specification requirement.**

Robustness is family-indexed: invariances track the experienced (ε). Engineering robust generalization therefore requires specifying (or learning) (ε) explicitly and evaluating accordingly (Prediction 3).

## 7.3 Physics and information theory: a cross-substrate language for prediction

**Loss-as-codelength enables substrate symmetry.**

Negative log-likelihood provides a common currency: codelength in AI [20,21]. We propose integrated mismatch/update cost as the biological analogue, for which signals like MMN [55] serve as candidate proxies. This convergence is physically mandated: any non-predictive information maintained by a system imposes a lower bound on energy dissipation [14]. Consequently, the exception tax paid by unstable shortcuts is not merely an informational penalty but a thermodynamic one.This enables cross-substrate comparison without relying on bit↔joule identifications.

**Rate–distortion as a feasible region.**

The framework does not assume systems sit exactly on the frontier; it predicts selection pressures constrain systems toward a feasible region near it, with drift off-frontier detectable via regret scaling and frontier-relative efficiency proxies (Section 6).

## 7.4 Limitations and scope conditions (explicit)

1. **Structured environments are required.** In "causal-poor" or effectively white-noise environments, expected log-loss is lower bounded by the entropy rate under correct modeling assumptions [33,35].
2. **Causality bridge is conditional.** The alignment between shift-stable invariants and causal mechanisms requires that environments provide intervention-rich shifts and that the underlying causal structure is approximately modular. These are empirical premises about the world, not derivable from information theory alone (Section 2.3).
3. **Nonparametric rates are allowed.** The key empirical signature is a separation between linear patch-driven growth and sublinear estimation-driven growth; the sublinear exponent may depend on effective dimension/manifold complexity.

4. **Implementation is not fixed.** Predictive coding is one plausible implementation; ITI/CEP is a constraint-and-selection account.
5. **Purely passive / non-intervenable environments.** In domains where intervention-like diversity is unavailable (e.g., forecasting processes that cannot be experimentally perturbed), CEP may select for shift-stable correlations rather than causal mechanisms. In such settings, the strongest causal claim does not apply; the relevant prediction is reduced excess codelength under the available shift family.

## 7.5 Open problems

**1. Formal regret separation theorems for representation learning under shifts.**

The toy separation in Section 2.2.3 provides an anchor case. A central open problem is deriving general conditions under which invariant/mechanism-aligned representations provably achieve sublinear cross-environment regret while shortcut/patch representations incur linear regret for broad model classes (building on standard online-learning regret theory [37,38]).

*2. Quantifying "shift richness" and the crossover E.**

A theoretical characterization of E* in terms of mechanism complexity, model class capacity, and shift-family entropy would make CEP sharply predictive and allow principled design of training curricula. A central practical implication is the finite-sample crossover: for small E (or T), a shortcut model with low slope can outperform a mechanism-aligned model with higher structural cost. Characterizing E* is therefore not only a theoretical goal but necessary for predicting when invariant/mechanism representations become "cheaper" than shortcuts under finite lifetimes, datasets, or compute budgets. This patching–invariance trade-off can be viewed as an MDL/universal-coding statement for multi-regime or nonstationary sources [21,30,40].

**3. Bridging biological signals to codelength with calibrated estimators.**

MMN and adaptation slopes are plausible proxies for integrated mismatch cost, but establishing calibrated mappings to predictive codelength (even up to monotone transforms) would strengthen cross-substrate comparability [68].

**4. Mechanism modularity in sensory and cognitive domains.**

The ICM bridge rests on approximate modularity. Empirically, it remains to delineate when (and at what granularity) biological domains satisfy modular-mechanism assumptions, and how active sensing scaffolds mechanism discovery.

# 8. Conclusion

This review addressed a concrete empirical puzzle: why do biological and artificial neural networks converge on similar representational solutions, and in key cases exhibit quantitative alignment, despite radically different origins, substrates, and optimization procedures? We proposed a constraint channel that makes this convergence expected under explicit conditions.

The **Information-Theoretic Imperative (ITI)** states that active persisters in uncertain environments are constrained toward predictive compression under finite resources. The **Compression Efficiency Principle (CEP)** specifies the mechanism under distribution shift: shortcut representations that exploit shift-unstable correlations pay an "exception tax" that appears formally as linear excess codelength (regret) in environmental diversity, whereas representations that encode shift-stable invariants amortize description length and incur sublinear regret dominated by estimation/approximation. Under ICM-rich interventional diversity and approximate modularity, these invariants align with causal-generative mechanisms in the operational sense relevant for intervention-invariant prediction and localized updating.

Together, ITI and CEP explain the three motivating biological phenomena—neural metabolic cost, near-efficient sensory coding, and hierarchical invariance—and connect them to parallel phenomena in modern deep learning: predictive loss as codelength, OOD brittleness as excess codelength under shift, and augmentation/interaction as engineered proxies for active sensing. The convergence of evolved and engineered systems is then not merely a parallel, but evidence for a substrate-independent constraint: efficient prediction under finite resources and shifting environments restricts viable representational strategies, channeling both systems toward similar solutions.

The framework is designed to be falsifiable. Section 6 specified protocols and explicit failure conditions—particularly regret-scaling separations under shortcut-flipping shifts and efficiency–robustness coupling within task families—aimed at generating decisive empirical traction across neuroscience and machine learning. If those predictions fail systematically under controlled conditions, ITI/CEP must be revised or rejected.

If they hold, the implications extend beyond explaining a curious empirical alignment. The convergence of brains and deep networks would then reflect something deeper: a basin of attraction in representational space that any sufficiently constrained prediction system must enter. Evolution found it through differential persistence over geological time. Gradient descent finds it through loss minimization over computational time. The paths differ; the destination converges—a result consistent with a physical-constraint perspective on intelligence.

Intelligence, on this view, is not a special achievement of carbon or silicon. It is what happens when matter is organized to persist under uncertainty: compression becomes necessary, shift-stability becomes advantageous, and mechanism-aligned representations become inevitable. The question is not whether such systems will discover causal structure, but how

long—measured in environments encountered or resources expended—before the crossing point is reached.

That deep networks trained on prediction converge with evolved neural systems is not merely an alignment curiosity—it suggests that we have identified, in information-theoretic and thermodynamic terms, physical principles sufficient to explain core aspects of biological intelligence. Just as statistical mechanics explains life's thermal behavior without specifying molecular detail, ITI/CEP proposes that compression under uncertainty explains representational architecture without specifying synaptic implementation.

# APPENDIX A. Formal definitions, notation, and core propositions

This appendix consolidates notation and the formal objects used throughout the manuscript. Proofs are sketched only where needed to fix scaling claims; the main text remains the authoritative narrative. Where we invoke standard regret and MDL results without full derivation, the usual regularity assumptions apply (e.g., stationarity within each environment e, a well-defined normalized likelihood $p_\theta$, finite effective dimension of the reference model class, and identifiable/invariant structure under the specified shift family).

## A.1 Symbol table

| Symbol | Meaning | Notes / scope |
|---|---|---|
| $e \in \varepsilon$ | environment / context / regime / intervention index | $\varepsilon$ is the shift family |
| $P_e(X,Y)$ | joint distribution in environment e | observational or interventional |
| $Q_e$ | shifted evaluation distribution for environment e | used in $L_{Q_e}$, $\Delta_e$ |
| $\mu$ | distribution over environments | used in $\Delta_\varepsilon = E_{e\sim\mu}[\Delta_e]$ |
| X | observations / sensory input | includes $X_{\le t}$ when indexed |
| Y | prediction target | label, future segment, or next outcome |
| Z | representation / latent code | produced by $q_\varphi(z\|x)$ |
| $q_\varphi(z\|x)$ | encoder | parameterized by $\varphi$; maps input to latent distribution |
| $p_\theta(y\|z)$ | predictive model | parameterized by $\theta$; maps latent code to prediction |
| $\ell = -\log p_\theta(y\|z)$ | per-sample negative log-likelihood | instantaneous loss; $E_P[\ell] = L_P(m)$ |
| $L_P(m)$ | expected codelength under P | $m=(\varphi,\theta)$ |
| $L_{Q_e}(m)$ | expected codelength under $Q_e$ | |
| Δ_e(m) | shift penalty | L_{Q_e}(m) - L_P(m) |
| Δ_ε(m) | aggregate shift penalty | E_{e~μ}[Δ_e(m)] |
| T | number of time steps / samples | used for online sequences of regimes |
| E | number of distinct environments encountered | alternative to time T |
| M | model class | representation + predictor family |
| Regret_T(M) | within-class sequential regret | defined in A.2 |
| M_short | shortcut/patch class | cannot represent invariants without patching |
| M_inv | invariant-capable class | can represent shift-stable structure |
| G_T | between-class codelength gap | defined in A.3 |
| θ_e | environment-specific correction/patch parameters | patch list element |
| c_patch(e) | incremental patch description length | c_patch(e) = L(θ_e) |
| C_patch(E) | cumulative patching cost | Σ_{e=1}^E c_patch(e) |
| c_min | lower bound on incremental patch cost | c_min = inf_e c_patch(e) (special case) |
| E* | crossover ("escape velocity") | defined via L_0 + C_patch(E) > L_inv (A.8) |
| L_0 | base model description length | environment-independent MDL baseline |
| L_inv | invariant/mechanism model description length | MDL structural cost |
| R(m) | resource measure | parameters/FLOPs/memory/energy proxy (within-substrate) |
| Êff | efficiency proxy | e.g., -ΔL/ΔR |
| S_t | latent state (agentic) | optional; generates observations/outcomes |
| A_t | action (agentic) | optional; chosen after Z_t |
| ε intervention-rich | shift richness condition | operationally defined in Section 2.3.2; formalized in A.6 |

## A.2 Predictive codelength and within-class regret

When $p_\theta(y|z)$ is normalized and $\ell = -\log p_\theta(y|z)$, cumulative loss is cumulative predictive codelength [20,21].

For sequence $(x_t,y_t)_{t=1}^T$, the regret of a sequential predictor $\hat{m}_t$ relative to the best fixed model in class M is [37]:

$$\mathbf{Regret_T(M) = \Sigma_{t=1}^T -\log p_{\hat{m}_{t-1}}(y_t|x_t) - \min_{m\in M} \Sigma_{t=1}^T -\log p_m(y_t|x_t).}$$

In parametric well-specified settings this is often $\tilde{O}(\log T)$; in nonparametric settings $\tilde{O}(\sqrt{T})$ or other sublinear rates are common [37]. ITI/CEP requires a separation from linear patch-driven growth under shift, not a specific sublinear exponent.

## A.3 Between-class gap (G_T)

$$\mathbf{G_T = \min_{m\in M_short} \Sigma_{t=1}^T \ell_t(m) - \min_{m\in M_inv} \Sigma_{t=1}^T \ell_t(m).}$$

Linear growth in $G_T$ under shift is the signature of shortcut mis-specification relative to the shift family; it is not a claim that "all regret is linear."

## A.4 Crossover point (E) (escape velocity)*

The crossover E* is the point at which accumulating environment-specific patches becomes more costly (in model description length) than representing invariant/mechanism structure.

Using the patching notation from Proposition A.2, define the cumulative patching cost

$$\mathbf{C_patch(E) \equiv \Sigma_{e=1}^E c_patch(e), \text{ where } c_patch(e) = L(\theta_e|\theta_{1:e-1}).}$$

Let $L_0$ be the environment-independent base model description length and $L_{inv}$ the description length of an invariant-capable model. (Under ICM conditions detailed in Section 2.3 and A.6, these invariants correspond to causal mechanisms.)

The general crossover condition is

**$L_0 + C_{patch}(E) > L_{inv}$.**

Accordingly, a natural operational definition is

*$E \equiv \inf\{E : L_0 + C_{patch}(E) > L_{inv}\}$.**

**Special case (algorithmically diverse shifts).** If $c_{patch}(e) \geq c_{min} > 0$ over the environments considered, then $C_{patch}(E) \geq E \cdot c_{min}$, yielding the sufficient condition

*$E \approx (L_{inv} - L_0) / c_{min}$.**

When shifts are low-dimensional and patches are jointly compressible, $C_{patch}(E)$ can grow sublinearly, delaying $E^*$ (see Proposition A.2).

**Units and substrate-independence.** In artificial systems, $L(\cdot)$ and $C_{patch}(E)$ are measured in bits of description length or excess codelength; in biological systems, the analogous patch cost can be operationalized via energetic costs of adaptation (integrated mismatch/update proxies). While units differ, $E^*$ and ratios such as $E/E^*$ remain dimensionless thresholds with the same meaning.

## A.5 Mutual information in deterministic representations

Mutual-information-based efficiency ratios (as used in some Information Bottleneck formulations) can be well-defined and useful under explicit stochastic encoders, discretization schemes, or noise models. However, for deterministic mappings $Z=f(X)$ with continuous $X$, $I(X;Z)$ can be ill-defined (divergent) without an explicit noise model; empirically, estimates can depend strongly on binning or injected noise, as emphasized in critiques of literal IB interpretations in deep networks (e.g., Saxe et al. [44]).

For this reason, the manuscript's empirical program is formulated in terms of **predictive codelength** (negative log-likelihood under normalized models) and **excess codelength (regret)** under shift, which are directly measurable and estimator-robust across both biological proxies and artificial systems.

## A.6 Intervention-rich shift families and ICM conditions

A shift family ($\varepsilon$) is **intervention-rich** when it contains regime changes or interventions that systematically vary different factors independently. This is an observational property: the shift family includes diverse perturbations that decorrelate potential confounds from stable predictive structure.

When the underlying causal structure exhibits approximate **modularity**—meaning the data-generating process can be decomposed into approximately autonomous mechanisms whose parameters can be changed independently—and when $\varepsilon$ provides interventions that perturb these mechanisms one at a time while leaving others approximately invariant, we say $\varepsilon$ satisfies the **Independent Causal Mechanisms (ICM) conditions** [28,29].

Whether a given shift family satisfies ICM conditions is an empirical question about environmental structure, not a theorem derivable from information theory. Operational criteria for determining when $\varepsilon$ satisfies ICM conditions are provided in Section 2.3.2, including:

1. Explicit experimental control over intervention targets
2. Natural experiments with identifiable causal structure
3. Posterior validation via localized adaptation patterns

Under ICM conditions, the shift-stable invariants selected by CEP (Section 2.2) correspond to causal mechanisms because mechanisms are precisely the representational units that (i) remain predictively valid across interventional shifts in $\varepsilon$, and (ii) require only localized updates when individual environmental factors change, thereby minimizing cross-environment codelength.

When $\varepsilon$ is intervention-rich but does not satisfy ICM conditions (e.g., due to entangled causal structure or purely observational shifts), CEP still selects shift-stable invariants, but these may correspond to stable confounders, symmetries, or other non-causal regularities rather than causal mechanisms (examples in Section 2.3.1).

## A.7 Proposition A.1

**Proposition A.1 (Linear loss gap under hidden shortcut flips).**

Let environments $e_t \in \{0,1\}$ recur with nonzero frequency and be unobserved. Let $Y=X_2$ in both environments and let shortcut feature $X_1$ flip ($X_1=Y$ if $e=0$, $X_1=1-Y$ if $e=1$). Let $M_{short}$ be predictors restricted to using $X_1$ and $M_{inv}$ predictors that can use $X_2$. Then for recurrent switching,

**$G_T = \Omega(T)$.**

**Proof sketch.** Any $X_1$-only predictor must incur a nonzero expected log-loss gap on a constant fraction of steps when the hidden context differs from its implicit mapping, yielding a per-step excess loss bounded below by a constant $c>0$. With recurrent switching, such steps occur $\Omega(T)$ times, giving total excess loss $\Omega(T)$. An $X_2$-based predictor is well-specified in both environments and pays only sublinear estimation regret.

## A.8 Proposition A.2 (patching lower bound) and crossover E*

Proposition A.1 shows that if a shortcut correlation flips across hidden contexts and the system does not represent the context, shortcut-only prediction incurs persistent excess loss and thus $\Omega(T)$ excess codelength. A complementary objection is that a sufficiently large system could "patch" each context. Proposition A.2 formalizes the cost of such patching under MDL/codelength accounting.

**Terminology note.** This appendix uses "invariant" to denote representations Z for which the predictive relationship $P(Y|Z)$ remains stable across environments ε, consistent with standard usage in machine learning [41,29]. The representation values themselves may change; what remains invariant is the conditional distribution. Under the additional ICM conditions detailed in Section 2.3 and A.6, these invariants correspond to causal mechanisms.

**Proposition A.2 (Patching lower bound under context diversity).**

Let $\{P_e(X,Y)\}_{e\in\varepsilon}$ *be a family of environments. Suppose there exists a shortcut representation X_short such that achieving near-optimal predictive codelength in each environment requires an environment-specific predictor* $p_e(y|X_{short})$. *Consider a patching strategy class M_patch that attains low loss by storing or learning environment-conditioned corrections* $\{\theta_e\}_{e=1}^E$, rather than capturing an invariant representation that shares parameters across environments.

Assume the shift family (ε) satisfies ICM conditions (Section 2.3, A.6) and the correction increments are approximately algorithmically independent across environments. This is a coding-theoretic expression of mechanism modularity: the additional description needed to specify the correction for environment e does not substantially reduce the description needed for other environments $e' \neq e$, meaning patches are not jointly compressible unless the model represents higher-level invariant structure.

Define the incremental patch description length

$$c_{patch}(e) \triangleq L(\theta_e \mid \theta_{1:e-1}),$$

and the cumulative patching cost

$$C_{patch}(E) \triangleq \Sigma_{e=1}^{E} c_{patch}(e).$$

Then any uniquely decodable code satisfies the lower bound

$$L(M_{patch}) \geq L_0 + C_{patch}(E),$$

where $L_0$ is the environment-independent base model cost.

**Special case (algorithmically diverse shifts).** If there exists c_min>0 such that c_patch(e)≥c_min for all environments considered (the "approximately algorithmically independent" regime), then

**L(M_patch) ≥ L_0 + E·c_min.**

If instead {θ_e} is compressible across environments (e.g., θ_e=f(e) for a low-complexity function f), then C_patch(E) can grow sublinearly for long ranges; in that case the patch list is not "pure patching" but is capturing invariant/context structure, moving toward M_inv.

*Corollary A.2.1 (Crossover / escape velocity E).**

Let M_inv be an invariant-capable strategy with description length L(M_inv)=L_inv that achieves comparable per-environment predictive codelength across ε without environment-specific patch lists.

The MDL preference switches from patching to invariants when

**L_0 + C_patch(E) > L_inv.**

In the special case c_patch(e)≥c_min>0, a sufficient crossover condition is L_0 + E·c_min > L_inv, yielding

*E ≡ (L_inv - L_0) / c_min.**

For E > E*, patching strategies satisfying the assumptions incur greater model description length than the invariant strategy and are disfavored under MDL-style total codelength minimization.

**Units and substrate-independence.** In artificial systems, L(·), c_patch(e), and C_patch(E) are measured in bits of description length or excess codelength. In biological systems, the analogous patch cost can be operationalized via energetic costs of adaptation (integrated mismatch/update cost proxies). While units differ across substrates, the crossover condition and ratios such as E/E* remain dimensionless and comparable in meaning.

**Remark (algorithmic view).** When causal structure is modular (independent mechanisms), mechanisms can be represented as short programs that generate many environments' distributions via modular parameter changes, whereas patching corresponds to a growing table of environment-specific corrections. Shift families satisfying ICM conditions (A.6) are precisely those where these corrections are not jointly compressible, making the mechanistic representation compression-efficient. Without modular causal structure, patches may be jointly compressible, and the invariant representation need not correspond to causal mechanisms.

**Remark (algorithmic diversity of shifts).** CEP's selection pressure is sharpest when (ε) has high effective diversity (large incremental c_patch(e)); when shifts are low-dimensional and patches are compressible, C_patch(E) can remain sublinear for long ranges, delaying E*.

**Related literature (nonstationary coding/regret).** The patching–invariance trade-off can be viewed as an MDL/universal-coding statement for multi-regime or nonstationary sources: when regimes require corrections that are not jointly compressible, total description length grows with the number/effective diversity of regimes (e.g., [21,30,40,70]).

# APPENDIX B. Empirical protocols and measurement details (experimental contract)

This appendix provides operational details for the core tests in Section 6.

## B.1 AI protocols: computing codelength, shift penalties, and regret scaling

### B.1.1 Predictive codelength and shift penalties

Report $L_P(m)$, $L_{Q_e}(m)$, $\Delta_e(m)$, and $\Delta_\varepsilon(m)$ with confidence intervals (bootstrap over samples; repeated seeds where feasible).

### B.1.2 Between-class gap ($G_E$) under multi-environment exposure

Define explicit classes $M_{short}$ and $M_{inv}$ and compute:

$$G_E = \min_{m \in M_{short}} \Sigma_{i=1}^{E} L_{Q_{e_i}}(m) - \min_{m \in M_{inv}} \Sigma_{i=1}^{E} L_{Q_{e_i}}(m),$$

fitting scaling only beyond the crossover $E^*$.

### B.1.3 Constructing shortcut-flipping shifts

Start with synthetic multi-environment generators (e.g., colored-MNIST style) where shortcut correlations flip across environments while invariant structure remains stable. Use natural-data suites (ImageNet-C, ObjectNet, WILDS) primarily to measure $\Delta_\varepsilon$.

### B.1.4 Power and reporting standards

For correlation tests, aim for $n \geq 50$ variants where feasible; otherwise report wide intervals and treat results as preliminary. Always report seeds, compute, and capacity matching.

## B.2 Biological protocols: integrated mismatch cost as regret proxy

### B.2.1 EEG/MEG: MMN integration window and scaling target

**Signal:** mismatch negativity (MMN) under predictive coding interpretations [55].

**Regret proxy definition (MMN area).** For environment $e_i$, define

$$M(e_i) \equiv \int_{t_1}^{t_2} MMN_{e_i}(t)\, dt,$$

with a pre-registered window $[t_1,t_2]$ (e.g., 100–250 ms, paradigm-dependent). Define cumulative mismatch cost:

$$M_{\le E} \equiv \Sigma_{i=1}^{E} M(e_i).$$

CEP predicts $M_{\le E}$ grows approximately linearly in a shortcut/patch regime and sublinearly in an invariant regime.

**What MMN is a proxy for (timescale reconciliation and cost type).** MMN is a transient error response, whereas regret/excess codelength is cumulative. Our proxy is therefore not an MMN peak on a single trial, but the **cumulative mismatch-processing load** over repeated exposures and across environments: $M_{\le E}$ is intended as an experimental analogue of the integrated cost of residual prediction errors.

Biologically, CEP's "exception tax" can be paid through two distinct physical channels:

(i) **signaling costs**: ongoing metabolic costs of transmitting prediction errors (action potentials, synaptic currents) when representations remain mismatched to the environment, and

(ii) **plasticity costs**: episodic metabolic costs of synaptic modification (LTP/LTD, protein synthesis, structural remodeling) required to reconfigure representations when incorporating patches or adapting to new regimes.

The MMN-area proxy $M_{\le E}$ primarily targets (i). To probe (ii), complementary readouts are needed (e.g., metabolic imaging during re-learning/adaptation, longer-timescale adaptation curves, and/or molecular/structural plasticity markers).

**Cue-flip paradigm.** Train a stable invariant rule while manipulating a shortcut cue whose validity flips across environments ($e=1,...,E$), holding the invariant rule fixed.

**Control for salience/complexity vs predictability.** To falsify the interpretation "MMN tracks shift-regret rather than stimulus salience," include a control in which physical deviance/complexity is matched but predictive value differs: hold the deviant magnitude and probability constant while manipulating whether the deviant violates a learned predictive regularity (predictable vs unpredictable regimes). If $M_{\le E}$ tracks only salience/complexity and not predictability under this control, the MMN-as-regret-proxy hypothesis is undermined.

#### B.2.2 fMRI: adaptation slopes

Use repetition suppression / adaptation slopes under systematic violations; integrate across environments analogously to $M_{\le E}$.

#### B.2.3 Electrophysiology

Use time-integrated spike counts or LFP power changes in populations implicated in mismatch/error signaling; compare scaling across E.

### B.3 Distinguishing patching vs relearning

Keep the invariant rule constant; manipulate only the shortcut cue validity; test transfer on stimuli where shortcut cues are absent/uninformative.

### B.4 MI measurement note (secondary)

Where MI is used, report estimator, calibration, and uncertainty; core claims should rely on codelength/regret.

# References

1. DiCarlo JJ, Zoccolan D, Rust NC. How Does the Brain Solve Visual Object Recognition? Neuron. 2012;73(3):415–434. doi:10.1016/j.neuron.2012.01.010

2. Pinto N, Doukhan D, DiCarlo JJ, Cox DD. A High-Throughput Screening Approach to Discovering Good Forms of Biologically Inspired Visual Representation. PLoS Comput Biol. 2009;5(11):e1000579. doi:10.1371/journal.pcbi.1000579

3. Schrimpf M, Kubilius J, Hong H, Majaj NJ, Rajalingham R, Issa EB, et al. Brain-Score: Which Artificial Neural Network for Object Recognition Is Most Brain-Like? bioRxiv [Preprint]. 2018. doi:10.1101/407007

4. Yamins DLK, Hong H, Cadieu CF, Solomon EA, Seibert D, DiCarlo JJ. Performance-Optimized Hierarchical Models Predict Neural Responses in Higher Visual Cortex. Proc Natl Acad Sci USA. 2014;111(23):8619–8624. doi:10.1073/pnas.1403112111

5. Kriegeskorte N. Deep Neural Networks: A New Framework for Modeling Biological Vision and Brain Information Processing. Annu Rev Vis Sci. 2015;1:417–446. doi:10.1146/annurev-vision-082114-035447

6. Barlow HB. Possible Principles Underlying the Transformations of Sensory Messages. In: Rosenblith WA, editor. Sensory Communication. Cambridge, MA: MIT Press; 1961. pp. 217–234.

7. Friston K. The Free-Energy Principle: A Unified Brain Theory? Nat Rev Neurosci. 2010;11(2):127–138. doi:10.1038/nrn2787

8. Kaplan J, McCandlish S, Henighan T, Brown TB, Chess B, Child R, et al. Scaling Laws for Neural Language Models. arXiv [Preprint]. 2020. doi:10.48550/arXiv.2001.08361

9. Hendrycks D, Dietterich T. Benchmarking Neural Network Robustness to Common Corruptions and Perturbations. Proc Int Conf Learn Represent (ICLR). 2019.

10. Laughlin SB. A Simple Coding Procedure Enhances a Neuron's Information Capacity. Z Naturforsch C. 1981;36(9–10):910–912. doi:10.1515/znc-1981-9-1040

11. Raichle ME, Mintun MA. Brain Work and Brain Imaging. Annu Rev Neurosci. 2006;29:449–476. doi:10.1146/annurev.neuro.29.051605.112819

12. Rust NC, DiCarlo JJ. Selectivity and Tolerance ("Invariance") Both Increase as Visual Information Propagates from Cortical Area V4 to IT. J Neurosci. 2010;30(39):12978–12995. doi:10.1523/JNEUROSCI.0179-10.2010

13. Harris JJ, Jolivet R, Attwell D. Synaptic Energy Use and Supply. Neuron. 2012;75(5):762–777. doi:10.1016/j.neuron.2012.08.019

14. Still S, Sivak DA, Bell AJ, Crooks GE. Thermodynamics of Prediction. Phys Rev Lett. 2012;109(12):120604. doi:10.1103/PhysRevLett.109.120604

15. Bengio Y, Lee DH, Bornschein J, Mesnard T, Lin Z. Towards Biologically Plausible Deep Learning. arXiv [Preprint]. 2015. doi:10.48550/arXiv.1502.04156

16. Lillicrap TP, Santoro A, Marris L, Akerman CJ, Hinton G. Backpropagation and the Brain. Nat Rev Neurosci. 2020;21(6):335–346. doi:10.1038/s41583-020-0277-3

17. Simoncelli EP, Olshausen BA. Natural Image Statistics and Neural Representation. Annu Rev Neurosci. 2001;24:1193–1216. doi:10.1146/annurev.neuro.24.1.1193

18. Yamins DLK, DiCarlo JJ. Using Goal-Driven Deep Learning Models to Understand Sensory Cortex. Nat Neurosci. 2016;19(3):356–365. doi:10.1038/nn.4244

19. Friston KJ, Stephan KE. Free-Energy and the Brain. Synthese. 2007;159(3):417–458. doi:10.1007/s11229-007-9237-y

20. Rissanen J. Modeling by Shortest Data Description. Automatica. 1978;14(5):465–471. doi:10.1016/0005-1098(78)90005-5

21. Grünwald PD. The Minimum Description Length Principle. Cambridge, MA: MIT Press; 2007. ISBN: 978-0-262-07281-6

22. Tishby N, Zaslavsky N. Deep Learning and the Information Bottleneck Principle. 2015 IEEE Information Theory Workshop (ITW). 2015. doi:10.1109/ITW.2015.7133169

23. Berger T. Rate Distortion Theory: A Mathematical Basis for Data Compression. Englewood Cliffs, NJ: Prentice-Hall; 1971.

24. Friston KJ, Kiebel S. Predictive Coding under the Free-Energy Principle. Philos Trans R Soc Lond B Biol Sci. 2009;364(1521):1211–1221. doi:10.1098/rstb.2008.0300

25. Schmidhuber J. Formal Theory of Creativity, Fun, and Intrinsic Motivation (1990–2010). IEEE Trans Auton Ment Dev. 2010;2(3):230–247. doi:10.1109/TAMD.2010.2056368

26. Schmidhuber J. A Possibility for Implementing Curiosity and Boredom in Model-Building Neural Controllers. In: Meyer JA, Wilson SW, editors. From Animals to Animats: Proc SAB'91. Cambridge, MA: MIT Press; 1991. pp. 222–227.

27. Ashby WR. An Introduction to Cybernetics. London: Chapman & Hall; 1956.

28. Schölkopf B, Locatello F, Bauer S, Ke NR, Kalchbrenner N, Goyal A, et al. Toward Causal Representation Learning. Proc IEEE. 2021;109(5):612–634. doi:10.1109/JPROC.2021.3058954

29. Peters J, Janzing D, Schölkopf B. Elements of Causal Inference: Foundations and Learning Algorithms. Cambridge, MA: MIT Press; 2017.

30. Merhav N, Feder M. Universal Prediction. IEEE Trans Inf Theory. 1998;44(6):2124–2147. doi:10.1109/18.720534

31. Attwell D, Laughlin SB. An Energy Budget for Signaling in the Grey Matter of the Brain. J Cereb Blood Flow Metab. 2001;21(10):1133–1145. doi:10.1097/00004647-200110000-00001

32. Lennie P. The Cost of Cortical Computation. Curr Biol. 2003;13(6):493–497. doi:10.1016/S0960-9822(03)00135-0

33. Cover TM, Thomas JA. Elements of Information Theory. 2nd ed. Hoboken, NJ: Wiley-Interscience; 2006. ISBN: 0-471-24195-4

34. Tishby N, Pereira FC, Bialek W. The Information Bottleneck Method. Proc 37th Annu Allerton Conf Commun Control Comput. 1999:368–377. doi:10.48550/arXiv.physics/0004057

35. Ziv J, Lempel A. A Universal Algorithm for Sequential Data Compression. IEEE Trans Inf Theory. 1977;23(3):337–343. doi:10.1109/TIT.1977.1055714

36. Conant RC, Ashby WR. Every Good Regulator of a System Must Be a Model of That System. Int J Syst Sci. 1970;1(2):89–97. doi:10.1080/00207727008920220

37. Cesa-Bianchi N, Lugosi G. Prediction, Learning, and Games. Cambridge, UK: Cambridge University Press; 2006.

38. Hazan E, Agarwal A, Kale S. Logarithmic Regret Algorithms for Online Convex Optimization. Mach Learn. 2007;69(2–3):169–192. doi:10.1007/s10994-007-5016-8

39. Friston KJ, FitzGerald T, Rigoli F, Schwartenbeck P, Pezzulo G. Active Inference: A Process Theory. Neural Comput. 2017;29(1):1–49. doi:10.1162/NECO_a_00912

40. Herbster M, Warmuth MK. Tracking the Best Expert. Mach Learn. 1998;32(2):151–178. doi:10.1023/A:1007424614876

41. Arjovsky M, Bottou L, Gulrajani I, Lopez-Paz D. Invariant Risk Minimization. arXiv [Preprint]. 2019. doi:10.48550/arXiv.1907.02893

42. Rao RPN, Ballard DH. Predictive Coding in the Visual Cortex: A Functional Interpretation of Some Extra-Classical Receptive-Field Effects. Nat Neurosci. 1999;2(1):79–87. doi:10.1038/4580

43. Shwartz-Ziv R, Tishby N. Opening the Black Box of Deep Neural Networks via Information. arXiv [Preprint]. 2017. doi:10.48550/arXiv.1703.00810

44. Saxe AM, Bansal Y, Dapello J, Advani M, Kolchinsky A, Tracey BD, et al. On the Information Bottleneck Theory of Deep Learning. J Stat Mech Theory Exp. 2019;2019(12):124020. doi:10.1088/1742-5468/ab3985

45. Kolchinsky A, Tracey BD, Wolpert DH. Nonlinear Information Bottleneck. Entropy. 2019;21(12):1181. doi:10.3390/e21121181

46. Goldfeld Z, Polyanskiy Y. The Information Bottleneck Problem and Its Applications in Machine Learning. IEEE J Sel Areas Inf Theory. 2020;1(1):19–38. doi:10.1109/JSAIT.2020.2991561

47. Chelombiev I, Houghton C, O'Donnell C. Adaptive Estimators Show Information Compression in Deep Neural Networks. arXiv [Preprint]. 2019. doi:10.48550/arXiv.1902.09037

48. Achille A, Soatto S. Information Dropout: Learning Optimal Representations Through Noisy Computation. IEEE Trans Pattern Anal Mach Intell. 2018;40(12):2897–2905. doi:10.48550/arXiv.1611.01353

49. Bialek W, Rieke F, de Ruyter van Steveninck RR, Warland D. Reading a Neural Code. Science. 1991;252(5014):1854–1857. doi:10.1126/science.2063199

50. Olshausen BA, Field DJ. Emergence of Simple-Cell Receptive Field Properties by Learning a Sparse Code for Natural Images. Nature. 1996;381(6583):607–609.

doi:10.1038/381607a0

51. Baddeley R, Abbott LF, Booth MCA, Sengpiel F, Freeman T, Wakeman EA, et al. Responses of Neurons in Primary and Inferior Temporal Visual Cortices to Natural Scenes. Proc R Soc Lond B Biol Sci. 1997;264(1389):1775–1783. doi:10.1098/rspb.1997.0246

52. Dayan P, Abbott LF. Theoretical Neuroscience: Computational and Mathematical Modeling of Neural Systems. Cambridge, MA: MIT Press; 2001.

53. Bajcsy R. Active Perception. Proc IEEE. 1988;76(8):966–1005. doi:10.1109/5.5968

54. Clark A. Whatever Next? Predictive Brains, Situated Agents, and the Future of Cognitive Science. Behav Brain Sci. 2013;36(3):181–204. doi:10.1017/S0140525X12000477

55. Garrido MI, Kilner JM, Stephan KE, Friston KJ. The Mismatch Negativity: A Review of Underlying Mechanisms. Clin Neurophysiol. 2009;120(3):453–463. doi:10.1016/j.clinph.2008.11.029

56. Shannon CE. A Mathematical Theory of Communication. Bell Syst Tech J. 1948;27(3):379–423. doi:10.1002/j.1538-7305.1948.tb01338.x

57. Krizhevsky A, Sutskever I, Hinton GE. ImageNet Classification with Deep Convolutional Neural Networks. Adv Neural Inf Process Syst. 2012;25:1097–1105.

58. He K, Zhang X, Ren S, Sun J. Deep Residual Learning for Image Recognition. Proc IEEE Conf Comput Vis Pattern Recognit. 2016:770–778. doi:10.1109/CVPR.2016.90

59. Chen T, Kornblith S, Norouzi M, Hinton G. A Simple Framework for Contrastive Learning of Visual Representations. Proc 37th Int Conf Mach Learn (ICML). 2020.

60. He K, Fan H, Wu Y, Xie S, Girshick R. Momentum Contrast for Unsupervised Visual Representation Learning. Proc IEEE/CVF Conf Comput Vis Pattern Recognit. 2020:9729–9738. doi:10.1109/CVPR42600.2020.00975

61. Barbu A, Mayo D, Alverio J, Luo W, Wang C, Gutfreund D, et al. ObjectNet: A Large-Scale Bias-Controlled Dataset for Pushing the Limits of Object Recognition Models. Adv Neural Inf Process Syst. 2019;32.

62. Koh PW, Sagawa S, Marklund H, Xie SM, Zhang M, Balsubramani A, et al. WILDS: A Benchmark of in-the-Wild Distribution Shifts. Proc 38th Int Conf Mach Learn (ICML). 2021.

63. Brown TB, Mann B, Ryder N, Subbiah M, Kaplan J, Dhariwal P, et al. Language Models Are Few-Shot Learners. Adv Neural Inf Process Syst. 2020;33:1877–1901.

64. Gao P, Trautmann E, Yu BM, Bhaskaran-Nair K, Shenoy K, Ganguli S, et al. A Theory of Multi-Neuron Dimensionality, Dynamics and Measurement. bioRxiv [Preprint]. 2017. doi:10.1101/214262

65. Belkin M, Hsu D, Ma S, Mandal S. Reconciling Modern Machine-Learning Practice and the Classical Bias–Variance Trade-Off. Proc Natl Acad Sci USA. 2019;116(32):15849–15854. doi:10.1073/pnas.1903070116

66. Wei J, Tay Y, Bommasani R, Raffel C, Zoph B, Borgeaud S, et al. Emergent Abilities of Large Language Models. Trans Mach Learn Res. 2022. doi:10.48550/arXiv.2206.07682

67. Akyürek E, Schuurmans D, Andreas J, Ma T, Zhou D. What Learning Algorithm Is In-Context Learning? Investigations with Linear Models. arXiv [Preprint]. 2022. doi:10.48550/arXiv.2211.15661

68. Grill-Spector K, Henson R, Martin A. Repetition and the Brain: Neural Models of Stimulus-Specific Effects. Trends Cogn Sci. 2006;10(1):14–23. doi:10.1016/j.tics.2005.11.006

69. Power A, Burda Y, Edwards H, Babuschkin I, Misra V. Grokking: Generalization Beyond Overfitting on Small Algorithmic Datasets. arXiv [Preprint]. 2022. doi:10.48550/arXiv.2201.02177

70. Delétang G, Ruoss A, Duquenne PA, Catt E, Genewein T, Mattern C, et al. Language Modeling Is Compression. Proc 12th Int Conf Learn Represent (ICLR). 2024. doi:10.48550/arXiv.2309.10668

71. Valmeekam CK, Narayanan K, Kalathil D, Chamberland JF, Shakkottai S. LLMZip: Lossless Text Compression Using Large Language Models. arXiv [Preprint]. 2023. doi:10.48550/arXiv.2306.04050